\documentclass{article}

\usepackage{arxiv}
\usepackage{amsmath}
\usepackage{amssymb}
\usepackage{booktabs}
\usepackage{multirow}
\usepackage[table,dvipsnames]{xcolor}
\usepackage{amsmath,amsfonts,amssymb, bm}
\usepackage{algorithmic}
\usepackage{array}
\usepackage{textcomp}
\usepackage{stfloats}
\usepackage{url}
\usepackage{verbatim}
\usepackage{graphicx}
\usepackage{cite}

\usepackage{times}
\usepackage{breqn}
\usepackage{comment}

\usepackage{framed,multirow}
\usepackage{caption}
\usepackage{subcaption}
\usepackage{booktabs}
\usepackage{soul}
\usepackage{bbding}

\usepackage{orcidlink}
\usepackage{lipsum}
\usepackage{hyperref}
\usepackage[utf8]{inputenc}
\usepackage{soulutf8}
\usepackage{pifont}
\usepackage{amsmath}
\usepackage{amssymb}
\usepackage{color,soul}
\usepackage[dvipsnames]{xcolor}
\usepackage{gensymb}
\usepackage{bm}
\usepackage{enumitem}

\newcommand{\mytilde}{\raise.17ex\hbox{$\scriptstyle\mathtt{\sim}$}}
\DeclareMathOperator*{\argmin}{arg\,min}

\let\emptyset\varnothing

\def\sss{\mathbf{s}}

\def\xx{\mathbf{x}}

\def\zz{\mathbf{z}}

\def\MM{\mathbf{M}}

\def\WW{\mathbf{W}}

\def\cC{\mathcal{C}}
\def\dD{\mathcal{D}}

\def\hH{\mathcal{H}}

\def\lL{\mathcal{L}}

\def\sS{\mathcal{S}}

\def\xX{\mathcal{X}}
\def\yY{\mathcal{Y}}

\def\Re{\mathbb{R}}

\newcommand{\resultnof}[2]{\ensuremath{#1}\scriptsize{$\pm$\ensuremath{#2}}}
\hypersetup{
  colorlinks   = true, 
  urlcolor     = blue, 
  linkcolor    = blue, 
  citecolor   = red 
}

\usepackage[utf8]{inputenc} 
\usepackage[T1]{fontenc}    
\usepackage{hyperref}       
\usepackage{url}            
\usepackage{booktabs}       
\usepackage{amsfonts}       
\usepackage{nicefrac}       
\usepackage{microtype}      
\usepackage{lipsum}
\usepackage{fancyhdr}       
\usepackage{graphicx}       
\graphicspath{{media/}}     

\title{Selective Attention-based Modulation for Continual Learning}

\author{
        \href{https://orcid.org/0000-0002-1333-8348}{\includegraphics[scale=0.06]{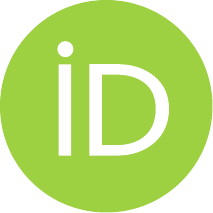}\hspace{1mm} Giovanni Bellitto}\\
        PeRCeiVe Lab \\
        University of Catania, Italy\\ 
        \And
        \href{https://orcid.org/0000-0002-6122-4249}{\includegraphics[scale=0.06]{orcid.pdf}\hspace{1mm}Federica Proietto Salanitri}\\
        PeRCeiVe Lab \\
        University of Catania, Italy\\ 
        \And
        \href{https://orcid.org/0000-0002-6721-4383}{\includegraphics[scale=0.06]{orcid.pdf}\hspace{1mm}Matteo Pennisi}\\
        PeRCeiVe Lab \\
        University of Catania, Italy\\ 
        \And
        \href{https://orcid.org/0000-0002-6721-4383}{\includegraphics[scale=0.06]{orcid.pdf}\hspace{1mm}Matteo Boschini}\\
        AImageLab -
        University of Modena\\ and Reggio Emilia, Italy\\ 
        \And
        \href{https://orcid.org/0000-0002-6721-4383}{\includegraphics[scale=0.06]{orcid.pdf}\hspace{1mm}Angelo Porrello}\\
        AImageLab -
        University of Modena\\ and Reggio Emilia, Italy\\ 
        \And
        \href{https://orcid.org/0000-0002-6721-4383}{\includegraphics[scale=0.06]{orcid.pdf}\hspace{1mm}Simone Calderara}\\
        AImageLab - 
        University of Modena\\ and Reggio Emilia, Italy\\ 
        \And
        \href{https://orcid.org/0000-0002-2441-0982}{\includegraphics[scale=0.06]{orcid.pdf}\hspace{1mm}Simone Palazzo}\\
        PeRCeiVe Lab \\
        University of Catania, Italy\\ 
        \And
        \href{https://orcid.org/0000-0001-6653-2577}{\includegraphics[scale=0.06]{orcid.pdf}\hspace{1mm}Concetto Spampinato}\\
	PeRCeiVe Lab \\
        University of Catania, Italy
}

\begin{document}

\maketitle

\begin{abstract}
We present \emph{SAM}, a biologically-plausible \emph{selective attention-driven modulation} approach to enhance classification models in a continual learning setting. Inspired by neurophysiological evidence that the primary visual cortex does not contribute to object manifold untangling for categorization and that primordial attention biases are still embedded in the modern brain, we propose to employ auxiliary saliency prediction features as a modulation signal to drive and stabilize the learning of a sequence of non-i.i.d. classification tasks. Experimental results confirm that SAM effectively enhances the performance (in some cases up to about twenty percent points) of state-of-the-art continual learning methods, both in class-incremental and task-incremental settings. Moreover, we show that attention-based modulation successfully encourages the learning of features that are more robust to the presence of spurious features and to adversarial attacks than baseline methods. Code is available at: \url{https://github.com/perceivelab/SAM}.

\end{abstract}

\section{Introduction}
\label{sec:intro}

\begin{figure}[ht!]
\vspace{5pt}
    \centering
    \includegraphics[width=0.6\columnwidth]{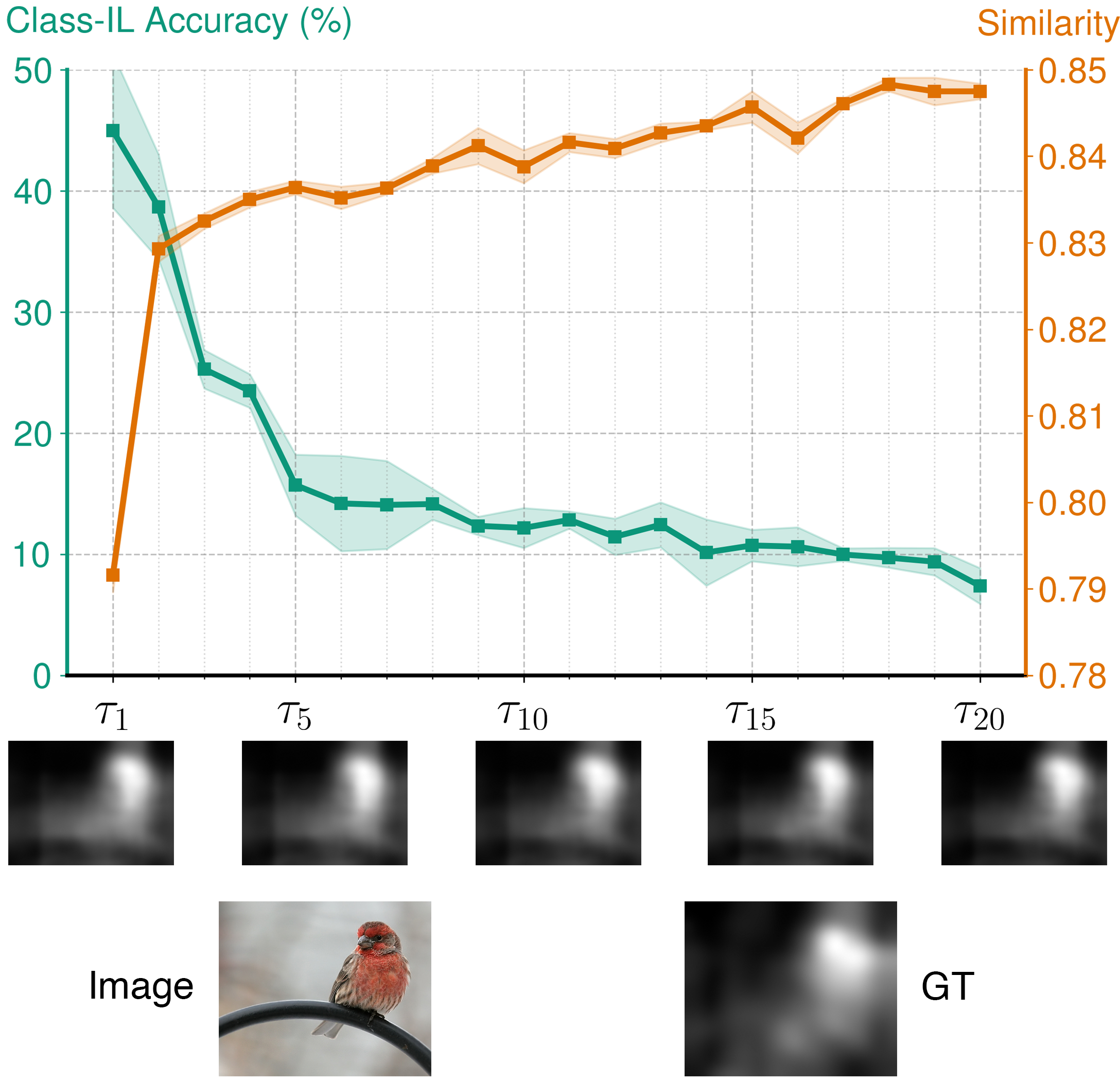}
    \caption{\textbf{Comparison between the forgetting-free behavior of saliency prediction and the typical catastrophic forgetting observed on classification tasks in continual learning scenarios}. Saliency accuracy (measured as \emph{similarity}~\cite{bylinskii2018different}) improves as the saliency network is presented with more tasks, while classification accuracy drops. This suggests that saliency detection is an i.i.d. task even in presence of a non-i.i.d. data distribution. Images on the $x$ axis show how predicted saliency maps are approximately constant over tasks.}
    \label{fig:iid-saliency}
\end{figure}

Humans possess the remarkable capability to keep learning, with limited forgetting of past experience, and to quickly re-adapt to new tasks and problems without disrupting consolidated knowledge.
Machine learning, on the contrary, has shown significant limitations when dealing with non-stationary data streams with a limited possibility to replay past examples. The main reason for this shortcoming can be found in the inherent structure, organization and optimization approaches of artificial neural networks, which differ significantly from how humans learn and how their neural connectivity is built when accumulating knowledge over a lifetime. According to the \emph{Complementary Learning Systems (CLS) theory}~\cite{pmid7624455,pmid27315762}, the human ability to learn effectively may be due to the interplay between two learning processes that originate, respectively, on the hippocampus and on the neocortex. These two brain regions interact to support learning representations from experience (the neocortex) while consolidating and sustaining long-term memory (the hippocampus). This theory has inspired several continual learning methods~\cite{pmid28292907,10.5555/3295222.3295393,kemker2017fearnet}. In particular, 
the recent DualNet method~\cite{pham2021dualnet} translates CLS concepts into a computational framework for continual learning. Specifically, it employs two learning networks: a \emph{slow learner}, emulating the memory consolidation process happening in the hippocampus through contrastive learning techniques, and a \emph{fast learner}, that aims at adapting current representations to new observations.
However, this strategy still appears insufficient for addressing the problem of continual learning, because it starts from the (possibly wrong) assumption that human neural networks directly process visual input with the objective of performing categorization from early vision layers. On the contrary, neurophysiological studies~\cite{pmid22325196,pmid17344377} are in near universal agreement that the object manifolds conveyed to primary visual cortex V1 (one of the earliest areas involved in vision) are as tangled as the pixel space. In other words, the neurons of the earliest vision areas do not contribute to object manifold untangling for categorization, but rather enforce luminance and contrast robustness~\cite{pmid17344377}.
This suggests that training early neurons with a visual categorization objective --- as done not only in DualNet, but in all existing continual learning methods --- is in stark contrast to the biological counterparts observed in primates.
Moreover, recent studies on the causes of forgetting in artificial neural networks showed that deeper layers (i.e., closer to the output) are less stable in presence of task shifts~\cite{ramasesh2021anatomy}, which is consistent with the hypothesis that earlier layers do not bear specific categorization responsibilities.

Given these premises, it is peculiar that existing bio-inspired continual learning methods tend to ignore all upstream neural processes underlying visual categorization, such as visual attention. Indeed, the ability to select relevant visual information appears to be the hallmark of human/primate cognition. Moreover, recent findings in cognitive neuroscience have shown that the visual attention priorities of human hunter-gatherer ancestors are still embedded in the modern brain~\cite{pnas.0703913104}: humans pay attention faster to animals than to vehicles, although we now see more vehicles than animals.
This primordial attention bias embedded in human brains suggests that the neuronal circuits of the ventral visual pathway are somehow inherited, as a form of genetic legacy from ancestral experience, and tend to remain stable over time --- thus not subject to forgetting, though we have long stopped hunting to survive.

Interestingly, we observed the same \textbf{forgetting-free} behavior for saliency prediction on artificial neural networks. Fig.~\ref{fig:iid-saliency} shows the trend of the \emph{similarity}~\cite{bylinskii2018different} metric for a saliency prediction model trained in a continual learning scenario, and compares it to the accuracy of a classification model under the same settings. While classification accuracy drops as the classifier learns new classes, the saliency metric remains stable, and even slightly improves.

From this observation, in this paper we propose \emph{SAM}, a \emph{Selective Attention-driven Modulation} strategy that employs saliency prediction~\cite{Borji2019} to drive the learning of a sequence of classification tasks in a continual learning setting. To emulate what has been observed in primates, where visual attention modulates the firing rate of neurons that represent the attended stimulus at different stages of visual processing~\cite{pmid10376597,pmid15120065}, SAM adopts a two-branch model: one branch performs visual saliency prediction~\cite{linardos2021deepgaze,jiang2015salicon,droste2020unified}, and its responses modulate (through multiplication) the features learned by a paired classification model in the second branch. 
SAM is model-agnostic and can be used in combination to any continual learning method. We demonstrate that saliency modulation positively impacts classification performance in online continual learning settings, leading to a significant gain in accuracy (up to 20 percent points) w.r.t. baseline methods. We further demonstrate the usefulness of saliency modulation on different benchmarks (including a challenging one that tackles fine-grained classification) and substantiate our claims through a set of ablation studies.
We finally show that saliency modulation, besides being biologically plausible, leads to learn saliency-modulated features that are more robust to the presence of spurious features and to adversarial attacks.


\section{Related Work}
Continual Learning (CL)~\cite{de2019continual,parisi2019continual} is a recently-popularized branch of machine learning whose objective is to bridge the gap in incremental learning between humans and neural networks. McCloskey and Cohen~\cite{mccloskey1989catastrophic} highlight that the latter experience a \emph{catastrophic forgetting} of previously acquired knowledge in the presence of distribution shifts in the input data stream. To compensate for this problem, countless solutions have been proposed that introduce either adequate regularization terms~\cite{kirkpatrick2017overcoming, zenke2017continual}, specific architectural organization~\cite{schwarz2018progress,mallya2018packnet} or the rehearsal of a small number of previously encountered data points~\cite{robins1995catastrophic,rebuffi2017icarl,buzzega2020dark}.

While current solutions help mitigating forgetting, their application to real-world settings proves difficult, as typical CL evaluations are conducted in accordance to unrealistic benchmarks~\cite{aljundi2019task,van2022three}. \emph{Online CL} (OCL)~\cite{mai2022online} addresses this issue by forbidding multiple epochs on the input stream. This is meant to model the realistic assumption that any data point captured in the wild occurs only once. 

To reach reasonable performance, most approaches tackling this challenging scenario adopt a replay strategy~\cite{ratcliff1990connectionist,robins1995catastrophic}. Some works focus on memory management: GSS~\cite{aljundi2019gradient} introduces a specific optimization of the basic rehearsal formula meant to store maximally informative samples in memory, while HAL~\cite{chaudhry2021using} individuates synthetic replay data points that are maximally affected by forgetting. Other approaches propose tailored classification schemes: CoPE~\cite{de2021continual} uses class prototypes to ensure a gradual evolution of the shared latent space; ER-ACE~\cite{caccia2022new} makes the cross-entropy loss asymmetric to minimize imbalance between current and past tasks. Finally, other works introduce a surrogate optimization objective: SCR~\cite{mai2021supervised} employs a supervised contrastive learning objective and OCM~\cite{guo2022online} leverages mutual information objectives: both aim at learning informative features that are less subject to forgetting.\\
Our proposal adopts a remarkably different approach w.r.t.\ these classes of methods, in that we take inspiration from cognitive neuroscience theory of learning and exploiting the features of a conjugate forgetting-free task (i.e., saliency prediction) to modulate the responses of our OCL model. Doing so produces a stabilizing effect on our model and makes it more resilient to forgetting. 

An approach that is similarly inspired by cognitive theories is DualNet~\cite{pham2021dualnet}, which employs two networks that loosely emulate how slow and fast learning work in humans. However, DualNet employs contrastive learning on the slow network (the earliest layers of the model), while it seems that object-identifying transformations happens later in the human visual system~\cite{pmid22325196,pmid17344377}. Our results, reported later, substantiate the suitability of our choice to use low-level processes, such as selective attention, to drive continual learning tasks, rather than contrastive learning or classification pre-training techniques as, respectively, in DualNet and TwF~\cite{boschini2022transfer}.
Finally, our work, like DualNet, follows the emerging \emph{NeuroAI}~\cite{neuroai} paradigm promoted by deep learning pioneers, according to which human neural computation will drive the next revolution in AI, bringing machines closer to human capabilities.

Despite the idea of using saliency prediction maps in continual learning has never been proposed, we have assisted to a recent trends where mitigated if the model is encouraged to recall the evidence for previously made decisions, stored as activation maps~\cite{ebrahimi2021remembering}. Specifically, it employs explainability techniques as Gradient-weighted Class Activation Mapping (Grad-CAM)~\cite{selvaraju2019grad} to store visual model explanations for each sample in the buffer and ensures model consistency with previous decisions during the training phase. Similarly, EPR~\cite{saha2023saliency}, instead of retaining whole images, employs Grad-CAM to identify the important patches and stores them in the episodic memory.\\
The above methods rely on using activation maps (sometime referred to as saliency maps) as regularizers to limit forgetting. However, it is worth to highlight that there exist a fundamental difference with our approach. Within the context of explainable artificial intelligence (XAI), techniques as Grad-CAM are utilized to generate \emph{attribution maps} to support a model prediction in term of relevant input features. They aim to identify important visual areas for a pre-trained classifier, a saliency predictor is a neural network trained with the aim to predict the area of a scene that will capture the attention of a human observer. Moreover, as reported in our experimental results, attribution maps tend to degrade over time, since they strictly depend on the internal state of a neural network, which is subject to forgetting.
In contrast, our saliency-based modulation, stemming directly from neuroscience theory and from our finding of forgetting-free behaviour of saliency is novel and unexplored.

\section{Method}
\label{sec:method}

\subsection{Online Continual Learning}

Following the recent literature, we pose OCL as a supervised image classification problem with an online non-i.i.d. stream of data, where each training sample is only seen once. Although our attention-driven modulation does not require the presence or knowledge of \emph{task boundaries}, in this formulation and in our experiments we assume that these are given, to the benefit of any baseline method enhanced by the proposed extension.

More formally, let $\dD =\{\dD_1, \dots, \dD_T\}$ be a sequence of data streams, where each pair $\left( \xx, y \right) \sim \dD_i$ denotes a data point $\xx \in \xX$ with the corresponding class label $y \in \yY$; the sample distributions (in terms of both the data point distribution and the class label distribution) of different $\dD_i$ and $\dD_j$ may vary --- for instance, class labels from $\dD_i$ might be different from those from $\dD_j$, though both must belong to the same domain $\yY$. 
Given a classifier $f: \xX \rightarrow \yY$, parameterized by $\bm{\theta}$,
the objective of OCL is to train $f$ on $\mathcal{D}$, organized as a sequence of $T$ tasks $\{\tau_1, \dots, \tau_T\}$, under the constraint that, at a generic task $\tau_i$, the model receives inputs sampled from the corresponding data distribution, i.e., $(\xx,y) \sim D_i$, and sees each sample only once during the whole training procedure. The classification model may optionally keep a limited \emph{memory buffer} $\MM$ of past samples, to reduce forgetting of features from previous tasks. The model update step between tasks can be summarized as:
\begin{equation}
    \langle f, \bm{\theta}_{i-1}, \dD_{i-1}, \MM_{{{i-1}}}\rangle \rightarrow \langle f, \bm{\theta}_{{i}}, \MM_{i}\rangle 
\end{equation}

where $\bm{\theta}_{i}$ and $\MM_{i}$ represent the set of model parameters and the buffer at the end of task $\tau_i$, respectively.
For methods that do not exploit buffer, $\MM_{i} = \emptyset, \forall{i}$. 

The training objective is to optimize a classification loss over the sequence of tasks (without losing accuracy on past tasks) by the model instance at the end of training:
\begin{dmath}
\argmin_{\bm{\theta}_T}
\sum_{i=1}^T
\mathbb{E}_{(\xx,y) \sim \dD_i}
\Bigl[
\lL \Bigl( f \left( \xx ; \bm{\theta}_T \right) , y \Bigr)
\Bigr]
\label{eq:cl_obj}
\end{dmath}
where $\lL$ is a generic classification loss (e.g., cross-entropy), which a continual learning model attempts to optimize while accounting for model \emph{plasticity} (the capability to learn current task data) and \emph{stability} (the capability to retain knowledge of previous tasks)~\cite{mccloskey1989catastrophic}. 

\subsection{SAM: Selective Attention-driven Modulation}
\begin{figure*}[t]
    \centering
    \includegraphics[width=\textwidth]{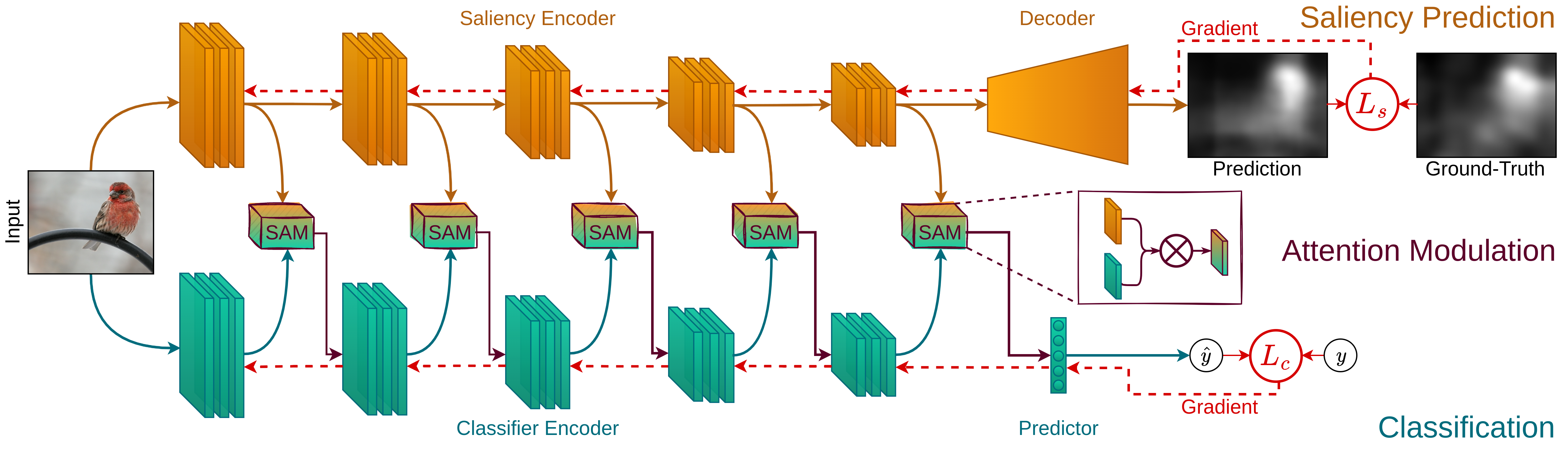}
    \caption{\textbf{Architecture of the proposed selective attention-based modulation (SAM) strategy. }The classification backbone is paired with a saliency prediction network that, given its capability of being forgetting-free, aims at adjusting the learned classification features in order to mitigate overall forgetting.}
    \label{fig:arch}
\end{figure*}

Our method is grounded on the neurophysiological evidence that attention-driven neuronal firing rate modulation is multiplicative and the scaling of neuronal responses depends on the similarity between a neuron's preferred stimulus and the attended feature~\cite{pmid10376597,pmid15120065}. This hypothesis is translated into a general artificial neural architecture, where we emulate the the process of human selective attention through a visual saliency prediction network\footnote{Presenting saliency prediction methods is out of the scope of this paper; an extensive survey can be found in~\cite{Borji2019}.} whose activations modulate, through multiplication, neuron activations of a paired classification network at different stages of visual processing.
Formally, let $S: \xX \rightarrow \sS$ be a saliency prediction network, where $\xX$ is the space of input images and $\sS$ the space of output saliency maps. Generally, if $\xX = \Re^{3\times H\times W}$ for RGB images, then $\sS=\Re^{H \times W}$, where each location of a map $\sss \in \sS$ measures the \emph{saliency} of the corresponding pixel in the RGB space. We assume that $S$ can be decomposed into two functions, an encoder $E: \xX \rightarrow \hH$ and a decoder $D: \hH \rightarrow \sS$, such that $S\left(\xx\right) = D\left(E\left(\xx\right)\right)$, for $\xx \in \xX$.
Then, given an online continual learning problem with data stream $\dD$ and set of classes $\yY$, let $C: \xX \rightarrow \yY$ be a classification network, such that $C$ and the saliency encoder $E$ share the same architecture (with independent parameters). An illustration of the proposed architecture is shown in Fig.~\ref{fig:arch}.\\
At training time, both $S$ and $C$ observe the same data stream, from which pairs $\left( \xx, y\right)$ of input data and class label are iteratively sampled. Through the use of an external \emph{saliency oracle}, we extend each data sample to a triple $\left( \xx, y, \sss \right)$, where $\sss$ is the target saliency map associated to $\xx$. The oracle can be either a set of ground-truth maps, when available, or \emph{pseudo-labels} provided as the output of a pre-trained saliency predictor (unrelated to $S$). We therefore proceed to optimize a multi-objective loss function $\lL = \lL_s + \lambda \lL_c$, with $\lambda$ being a weighing hyperparameter. Loss term $\lL_s$ is computed on the output of saliency predictor $S$, and compares the estimated saliency map $S(\xx)$ with the target $\sss$ by means of the Kullback-Leibler divergence (commonly employed as a saliency prediction objective~\cite{bylinskii2018different, droste2020unified, bellitto2021hierarchical, wang2021spatio, Hu_2023_WACV}):
\begin{equation}
\lL_s = \sum_i s_i \log \left( \frac{s_i}{S_i(\xx) + \epsilon}  + \epsilon \right)
\label{eq:l_s}
\end{equation}
with $s_i$ and $S_i(\xx)$ iterating over map pixels in $\sss$ and $S(\xx)$, respectively.
Loss term $\lL_c$ encodes a generic online continual learning objective, as introduced in Eq.~\ref{eq:cl_obj}. As the proposed approach is method-agnostic, details on the formulation of $\lL_c$ may vary. 

In order to enforce selective attention-driven modulation of classification neuronal activations, we leverage the architectural identity of saliency prediction encoder $E$ and classifier $C$ to alter the feedforward pass of the latter, by multiplying pre-activation features in $C$ by the corresponding features in $E$, before applying a non-linearity and feeding them to the next layer of the network. Formally, let us assume that the $C$ and $E$ networks consist of a sequence of layers $\left\{ l_1, l_2, \dots, l_L\right\}$. Without loss of generality, let each layer $l_i$ compute its output as $\zz_i = \sigma \left( \WW_i \zz_{i-1}\right)$, with $\sigma$ being an activation function, $\WW_i$ the network-specific layer parameters (i.e., not shared between $E$ and $C$) and $\zz_{i-1}$ the output of the previous layer (or the network's input $\xx$, if appropriate). Then, let us distinguish between features $\zz_{i}^{(s)}$ and $\zz_{i}^{(c)}$, respectively representing the output of layer $l_i$ by the saliency prediction encoder $S$ and the classifier $C$. We apply attention-driven modulation by modifying the computation of $\zz_{i}^{(c)}$ as follows:
\begin{equation}
\zz_{i}^{(c)} = \sigma\biggl( \WW_i^{(c)} \left( \zz_{i-1}^{(c)} \odot \zz_{i-1}^{(s)} \right) \biggr)
\label{eq:sam}
\end{equation}
where $\odot$ denotes the Hadamard product.
Intuitively, the proposed approach encourages the classification model to attend to ``salient'' features of the input, where the concept of \emph{saliency} is generalized from the pixel space to hidden representations. It is important to note that, at training time, gradient descent optimization of $\lL_c$ would also affect on the saliency encoder $E$. This is undesirable, as we previously showed (see Fig.~\ref{fig:iid-saliency}) that saliency features are robust to task shifts, unlike classification features: hence, in order to guarantee this property, we stop the gradient flow from $\lL_c$ to parameters in $E$, and use it to update the parameters of classifier $C$ only.

In the above formulation, we assumed the presence of a classification network with fully-connected layers; however, our method can be applied in an agnostic manner to any method employing, at least in part, a feature extractor implemented as a neural network. As such, the proposed method can be equally applied, for instance, both to end-to-end classification models (e.g., DER++~\cite{buzzega2020dark}) and to approaches with a neural backbone that computes class-representative prototypes (e.g., CoPE~\cite{de2021continual}).

\section{Experimental Results}
\subsection{Benchmarks}
We build two OCL benchmarks by taking image classification datasets and splitting their classes equally into a series of disjoint tasks:
\begin{itemize}[noitemsep,nolistsep,leftmargin=*]
    \item \textbf{Split Mini-ImageNet}~\cite{chaudhry2019tiny,ebrahimi2020adversarial,derakhshani2021kernel} is a challenging dataset that includes 100 classes from ImageNet, allowing for a longer task sequence. For each class, 500 images are used for training and 100 for evaluation. 
    \item \textbf{Split FG-ImageNet}\footnote{Split\ FG-ImageNet is derived from \url{https://www.kaggle.com/datasets/ambityga/imagenet100}} is a benchmark for fine-grained image classification that we use to test CL methods on a more challenging task than traditional ones.
    It includes 100 classes of animals extracted from ImageNet, belonging to 7 different species (\emph{annelids}, \emph{arachnids}, \emph{birds}, \emph{clams}, \emph{fishes}, \emph{reptiles}, \emph{shellfish}), reducing inter-class variability and leading to harder tasks. Each class contains 500 samples for training and 50 for evaluation.  
\end{itemize}

For both datasets, images are resized to 288$\times$384 pixels and split into twenty 5-way classification tasks.

\subsection{Training and Evaluation Procedure}
\label{sec:training}

\noindent \textbf{Baseline methods.} We evaluate the contribution of the SAM strategy when paired to a classification network trained using several state-of-the-art continual learning approaches, including rehearsal and non-rehearsal methods:
\begin{itemize}[nolistsep,noitemsep,leftmargin=*]
    \item \textbf{DER++}~\cite{boschini2022class}: a seminal work that combines rehearsal and knowledge distillation strategies for supporting model plasticity while limiting forgetting.
    \item \textbf{ER-ACE}~\cite{caccia2022new}: a variant of experience replay~\cite{ratcliff1990connectionist,robins1995catastrophic} which aims to prevent imbalances due to the simultaneous optimization of the current and past tasks by selectively masking softmax outputs.
    \item \textbf{CoPE}~\cite{de2021continual}: a prototype-based classifier with experience replay, whose careful update scheme prevents sudden disruptions in the latent space during incremental learning.  
    \item \textbf{LwF}~\cite{li2017learning}: a non-rehearsal method that enforces a model to preserve outputs of past model instances on new samples to limit forgetting.
    \item \textbf{oEWC}~\cite{ewc}: a non-rehearsal method that mitigates forgetting by selectively limiting the changes on weights that are most informative of past tasks.
\end{itemize}

All above methods employ ResNet-18~\cite{he2016deep} as a feature extraction backbone. We also report the results of jointly training the model on all classes for one epoch (\textbf{Joint}), and of training sequentially on each task without any particular countermeasure for avoiding forgetting (\textbf{Fine-tune}).

\noindent\textbf{Implementation details.} We apply the SAM strategy at five feature modulation points of ResNet-18's architecture, namely, the outputs of the first convolutional block and of the four main residual blocks. In compliance with online learning, all models are trained for a single epoch, using SGD as optimizer, with a fixed batch size of 8 both for the input stream and the replay buffer. 
Rehearsal methods are evaluated with three different sizes of the memory buffer (1000, 2000 and 5000). When applying SAM, besides each method's specific training objective, we also optimize the saliency prediction loss $\lL_s$ from Eq.~\ref{eq:l_s}. Saliency is estimated using DeepGaze IIE network~\cite{linardos2021deepgaze} as oracle. 

When using SAM, classifier $C$ and saliency predictor $S$ are identical ResNet-18 architectures, followed --- respectively --- by a linear classification layer and a saliency map decoder. The saliency decoder is broadly inspired by UNISAL~\cite{droste2020unified}, because of the low number of parameter it requires, which leads to a short runtime if compared with other saliency models.
While $C$ is trained from scratch, we employ a pre-trained saliency predictor $S$, consistently with neuroscience evidence showing that humans have selective attention already embedded in the brain~\cite{pnas.0703913104}. For a fair comparison, in all our experiments feature extraction backbones of baseline methods are initialized to the same pre-trained weights as $S$ (except where explicitly stated).
Care was taken to ensure that the set of OCL classes $\cC$ did not semantically overlap with pre-training data, to prevent any contamination from the saliency predictor to the classification task. 
Specifically, $S$ was pre-trained for 20 epochs on a subset of 100 ImageNet classes (disjoint from our two main benchmark datasets), using DeepGaze IIE as oracle. No class label information was used at this stage.

\noindent\textbf{Metrics and evaluation}. As a primary metric of OCL model performance, we report the \emph{final average accuracy} as $\frac{1}{T}\sum_{i=1}^{T}{a_i^T}$, 
where $a_i^T$ is the accuracy of the final model 
on the test set of task $\tau_i$. Accuracy $a_i^T$ can be computed in a \emph{Class-Incremental Learning} (\emph{Class-IL}) or in a \emph{Task-Incremental Learning} (\emph{Task-IL}) setting. In the latter, we assume that a task identifier is provided to the model at inference time, simplifying the problem by restricting the set of class predictions for a given sample. 
While task-incremental learning is often depicted as a trivial scenario in recent literature~\cite{farquhar2018towards,van2019three,aljundi2019gradient}, we emphasize its usefulness, as it isolates the effect of within-task forgetting from the model's bias towards the currently learned classes~\cite{wu2019large,hou2019learning,boschini2022class}. For this reason, we report both Class-Incremental and Task-Incremental performance in the results. Results are reported in terms of mean and standard deviation over five different runs.
\subsection{Results}
We first evaluate the contribution that attention-driven modulation provides to state-of-the-art OCL baselines. For each method, we compute Class-Incremental and Task-Incremental accuracy and compare to those obtained
when integrating SAM, as described in Sect.~\ref{sec:method}.
Results for rehearsal methods are reported in Table~\ref{tab:perf_online1epoch_class}, showing a pattern of enhanced performance when integrating SAM, for all tested buffer sizes. 
\begin{table*}[ht!]
\caption{\textbf{Class-Incremental and Task-Incremental accuracy of rehearsal methods} with and without SAM.}
\centering
\renewcommand{\arraystretch}{1.2}
\normalsize{
\begin{tabular}{l|ccc|ccc} 
\toprule
\textbf{Model}          & \multicolumn{3}{c}{\textbf{Split Mini-ImageNet}}                                                        & \multicolumn{3}{c}{\textbf{Split FG-ImageNet}}\\ 
\midrule
\multicolumn{7}{c}{\emph{Class-Incremental Learning}} \\
\midrule
Joint                                           & \multicolumn{3}{c}{\resultnof{14.79}{1.17}}                                    & \multicolumn{3}{c}{\resultnof{9.06}{1.07}}\\
\rowcolor{gray!10}
\hspace{0.2 cm }\textbf{$\hookrightarrow$SAM}   & \multicolumn{3}{c}{\resultnof{\textbf{16.10}}{0.30}}                           & \multicolumn{3}{c}{\resultnof{\textbf{9.73}}{0.73}}\\
Fine-tune                                       & \multicolumn{3}{c}{\resultnof{3.43}{0.35}}                                     & \multicolumn{3}{c}{\resultnof{2.43}{0.81}}\\
\rowcolor{gray!10}
\hspace{0.2 cm }\textbf{$\hookrightarrow$SAM}   & \multicolumn{3}{c}{\resultnof{\textbf{4.20}}{0.27}}                             & \multicolumn{3}{c}{\resultnof{\textbf{3.68}}{0.44}}\\
\midrule
\textbf{Buffer size}                            &\textbf{1000}                          & \textbf{2000}                          & \textbf{5000}                          &\textbf{1000}                            & \textbf{2000}                          & \textbf{5000}                        \\ 
\midrule
\arrayrulecolor{black}
DER++                                           & \resultnof{14.95}{3.11}           & \resultnof{12.82}{4.97}            & \resultnof{14.58}{2.55}            & \resultnof{8.08}{1.54}              & \resultnof{8.27}{1.72}             & \resultnof{9.20}{0.86}    \\
\rowcolor{gray!10}
\hspace{0.2 cm }\textbf{$\hookrightarrow$SAM}   & \resultnof{\textbf{19.13}}{1.62}  & \resultnof{{\textbf{22.92}}}{2.25} & \resultnof{{\textbf{25.35}}}{2.56} & \resultnof{\textbf{11.71}}{2.36}    & \resultnof{\textbf{12.97}}{1.62}   & \resultnof{\textbf{13.73}}{1.95} \\
ER-ACE                                          & \resultnof{20.86}{3.69}           & \resultnof{24.93}{3.20}            & \resultnof{26.31}{5.22}            & \resultnof{14.28}{0.96}             & \resultnof{16.45}{1.24}            & \resultnof{18.21}{3.45}   \\
\rowcolor{gray!10}
\hspace{0.2 cm }\textbf{$\hookrightarrow$SAM}   & \resultnof{\textbf{27.48}}{2.83}  & \resultnof{\textbf{33.09}}{1.28}   & \resultnof{\textbf{35.58}}{1.79}   & \resultnof{{\textbf{20.03}}}{3.13}  & \resultnof{\textbf{23.80}}{2.11}   & \resultnof{\textbf{28.68}}{0.50} \\
CoPE                                            & \resultnof{21.58}{1.60}           & \resultnof{23.58}{4.39}            & \resultnof{24.77}{3.56}            & \resultnof{16.45}{1.38}             & \resultnof{16.81}{0.83}            & \resultnof{17.77}{2.02}   \\
\rowcolor{gray!10}
\hspace{0.2 cm }\textbf{$\hookrightarrow$SAM}   & \resultnof{\textbf{26.66}}{2.22}  & \resultnof{\textbf{33.35}}{4.67}   & \resultnof{\textbf{45.04}}{2.44}   & \resultnof{\textbf{18.17}}{2.79}    & \resultnof{\textbf{27.14}}{1.62}   & \resultnof{\textbf{34.34}}{3.51} \\
\midrule
\multicolumn{7}{c}{\emph{Task-Incremental Learning}} \\
\midrule
Joint                                           & \multicolumn{3}{c}{\resultnof{63.12}{1.19}}                                       &  \multicolumn{3}{c}{\resultnof{56.33}{2.51}}\\
\rowcolor{gray!10}
\hspace{0.2 cm }\textbf{$\hookrightarrow$SAM}   & \multicolumn{3}{c}{\resultnof{\textbf{64.18}}{0.60}}                              &  \multicolumn{3}{c}{\resultnof{\textbf{56.72}}{1.09}}\\
Fine-tune                                       & \multicolumn{3}{c}{\resultnof{34.08}{2.28}}                                       &  \multicolumn{3}{c}{\resultnof{28.81}{1.66}}\\
\rowcolor{gray!10}
\hspace{0.2 cm }\textbf{$\hookrightarrow$SAM}   & \multicolumn{3}{c}{\resultnof{\textbf{57.07}}{3.44}}                              &  \multicolumn{3}{c}{\resultnof{\textbf{51.24}}{2.36}}\\
\midrule
\arrayrulecolor{black}
DER++                                           & \resultnof{73.07}{3.07}           & \resultnof{75.11}{5.61}              & \resultnof{77.71}{3.04}             & \resultnof{68.65}{2.14}              & \resultnof{70.24}{3.97}             & \resultnof{74.74}{1.14}    \\
\rowcolor{gray!10}
\hspace{0.2 cm }\textbf{$\hookrightarrow$SAM}   & \resultnof{\textbf{79.75}}{1.56}  & \resultnof{{\textbf{82.97}}}{0.25}   & \resultnof{{\textbf{84.10}}}{0.81}  & \resultnof{\textbf{72.83}}{3.90}     & \resultnof{\textbf{75.40}}{2.29}    & \resultnof{\textbf{78.26}}{1.10}  \\
ER-ACE                                          & \resultnof{71.00}{3.21}           & \resultnof{75.60}{3.47}              & \resultnof{77.17}{4.08}             & \resultnof{66.27}{0.92}              & \resultnof{69.09}{3.15}             & \resultnof{70.88}{5.72}    \\
\rowcolor{gray!10}
\hspace{0.2 cm }\textbf{$\hookrightarrow$SAM}   & \resultnof{\textbf{77.51}}{2.72}  & \resultnof{\textbf{82.22}}{0.96}     & \resultnof{\textbf{83.56}}{1.55}    & \resultnof{{\textbf{73.08}}}{2.14}   & \resultnof{\textbf{75.60}}{2.28}    & \resultnof{\textbf{79.46}}{0.56}  \\
CoPE                                            & \resultnof{68.00}{0.73}           & \resultnof{71.76}{2.95}              & \resultnof{74.31}{2.25}             & \resultnof{63.77}{2.32}              & \resultnof{67.29}{3.33}             & \resultnof{69.14}{2.93}    \\
\rowcolor{gray!10}
\hspace{0.2 cm }\textbf{$\hookrightarrow$SAM}   & \resultnof{\textbf{72.69}}{0.80} & \resultnof{\textbf{77.57}}{1.57}     & \resultnof{\textbf{84.64}}{1.20}    & \resultnof{\textbf{64.79}}{1.60}      & \resultnof{\textbf{73.39}}{1.11}    & \resultnof{\textbf{78.66}}{1.59}  \\
\bottomrule
\end{tabular}}

\label{tab:perf_online1epoch_class}
\end{table*}

\begin{table*}[ht]
\centering
\caption{\textbf{Class-Incremental and Task-Incremental accuracy of non-rehearsal methods} with and without SAM.}
\rowcolors{3}{white}{gray!10}
\renewcommand{\arraystretch}{1.2}
\normalsize{
\begin{tabular}{l|cc|cc} 
\toprule
\multirow{2}{*}{\textbf{Model}}                &\multicolumn{2}{c}{\textbf{Split Mini-ImageNet}}                             &\multicolumn{2}{c}{\textbf{Split FG-ImageNet}}\\ 
                                               &  Class-IL                            & Task-IL                             & Class-IL                          & Task-IL\\
\midrule
Joint                                          & \resultnof{14.79}{1.17}              & \resultnof{63.12}{1.19}             & \resultnof{9.06}{1.07}            &  \resultnof{56.33}{2.51}   \\
\hspace{0.1 cm }\textbf{$\hookrightarrow$SAM}  & \resultnof{\textbf{16.26}}{0.30}     & \resultnof{\textbf{64.34}}{0.59}    & \resultnof{\textbf{9.51}}{0.93}   &  \resultnof{\textbf{56.72}}{1.09}\\ 
Fine-tune                                      & \resultnof{3.43}{0.35}               & \resultnof{34.08}{2.28}             & \resultnof{2.43}{0.81}            &  \resultnof{28.81}{1.66}\\
\hspace{0.1 cm }\textbf{$\hookrightarrow$SAM}  & \resultnof{\textbf{4.20}}{0.27}      & \resultnof{\textbf{57.07}}{3.44}    & \resultnof{\textbf{3.68}}{0.44}   &  \resultnof{\textbf{51.24}}{2.36}\\
\midrule
LwF                                            & \resultnof{3.18}{0.41}               &  \resultnof{30.61}{1.80}            & \resultnof{3.25}{0.45}            & \resultnof{27.55}{1.64}         \\ 
\hspace{0.1 cm }\textbf{$\hookrightarrow$SAM}  & \resultnof{\textbf{4.22}}{0.31}      &  \resultnof{\textbf{48.61}}{2.14}   & \resultnof{\textbf{3.57}}{0.23}   & \resultnof{\textbf{36.57}}{2.09}\\ 
oEwC                                           & \resultnof{2.68}{0.24}               &  \resultnof{24.10}{1.55}            & \resultnof{2.38}{0.23}            & \resultnof{24.98}{1.15}         \\ 
\hspace{0.1 cm }\textbf{$\hookrightarrow$SAM}  & \resultnof{\textbf{3.08}}{0.31}      &  \resultnof{\textbf{35.33}}{3.18}   & \resultnof{\textbf{2.55}}{0.55}   & \resultnof{\textbf{26.02}}{1.64}\\ 

\bottomrule
\end{tabular}
}

\label{tab:perf_no_buffer}
\end{table*}

Table~\ref{tab:perf_no_buffer} shows results for non-rehearsal methods. In this case, SAM improvements are more evident in Task-Incremental; a marginal gain in Class-Incremental accuracy is also noticeable, though the low performance of baselines limits the room for improvements.

Since our strategy foresees two paired networks for classification and saliency prediction, we also compare with similar multi-branch CL baselines: 
\begin{itemize}[noitemsep,nolistsep,leftmargin=*]
    \item \textbf{DualNet}~\cite{pham2021dualnet}, mentioned in Sect.~\ref{sec:intro}, employs a dual-backbone architecture to decouple incremental classification (by a \emph{fast learner}) from self-supervised representation learning~\cite{zbontar2021barlow} (by a \emph{slow learner}).
    We adapt SAM to DualNet by replacing the slow learner and its training objective with our saliency prediction backbone, forcing the fast learner to use saliency features for classification.
    \item \textbf{TwF}~\cite{boschini2022transfer} employs a frozen pre-trained classification backbone to stabilize the learning of Class-Incremental features, by means of an attention mechanism. 
    To enable SAM, the pre-trained classification backbone and the feature distillation strategy are replaced with the saliency encoder, and the features of the two backbones are combined through multiplication, as described in Sect.~\ref{sec:method}.
\end{itemize}

\noindent Table~\ref{tab:perf_competitor} shows results for different buffer sizes\footnote{We could not run TwF with buffer size of 5000, due to excessive computing requirements.}. Integrating SAM outperforms baseline versions of both methods, suggesting that controlling learning through visual attention leads to better representation for classification than, for instance, contrastive learning. This is inline with cognitive neuroscience findings~\cite{pmid22325196,pmid19439676}, for which object identity-preservation, that also involves contrastive learning, happens mostly at later layers (e.g., IT neurons), while selective attention acts during the whole categorization process.

\begin{table*}[ht]
\centering
\caption{\textbf{Comparison of our saliency-attention mechanism to computational attention mechanisms (TwF~\cite{boschini2022transfer}) and contrastive learning (DualNet~\cite{pham2021dualnet})} for stabilizing learned classification features in CL tasks.} 
\rowcolors{3}{white}{gray!10}
\renewcommand{\arraystretch}{1.2}
\normalsize{
\begin{tabular}{c|l|cc|cc} 
\toprule
\multirow{2}{*}{\rotatebox[origin=c]{90}{\textbf{Buffer}}}  & \multirow{2}{*}{\textbf{Model}} &\multicolumn{2}{c}{\textbf{Split Mini-ImageNet}}&\multicolumn{2}{c}{\textbf{Split FG-ImageNet}}\\ 
&  & Class-IL & Task-IL & Class-IL & Task-IL\\
\midrule
\arrayrulecolor{black}
                                                & TwF                                           & \resultnof{23.78}{1.67}           & \resultnof{73.57}{1.27}           & \resultnof{15.32}{2.59}           & \resultnof{64.32}{5.18}\\
                                                & \hspace{0.2 cm }\textbf{$\hookrightarrow$SAM} & \resultnof{\textbf{28.36}}{3.72}  & \resultnof{\textbf{79.28}}{2.24}  & \resultnof{\textbf{20.04}}{1.63}  & \resultnof{\textbf{71.35}}{1.70}\\ 
                                                & DualNet                                       & \resultnof{20.57}{0.91}           & \resultnof{72.65}{0.56}           & \resultnof{15.62}{1.54}           & \resultnof{67.60}{1.56}\\
\multirow{-4}{*}{\rotatebox[origin=c]{90}{1000}}& \hspace{0.2 cm }\textbf{$\hookrightarrow$SAM} & \resultnof{\textbf{28.58}}{1.40}  & \resultnof{\textbf{81.79}}{0.59}  & \resultnof{\textbf{19.48}}{0.59}  & \resultnof{\textbf{75.76}}{0.51}\\
\midrule
                                                & TwF                                           & \resultnof{29.05}{2.02}           & \resultnof{78.38}{1.66}           & \resultnof{18.72}{1.75}           & \resultnof{72.15}{2.82}\\
                                                & \hspace{0.2 cm }\textbf{$\hookrightarrow$SAM} & \resultnof{\textbf{35.55}}{0.61}  & \resultnof{\textbf{82.98}}{0.85}  & \resultnof{\textbf{22.54}}{2.20}  & \resultnof{\textbf{73.34}}{2.94}\\
                                                & DualNet                                       & \resultnof{27.41}{1.79}           & \resultnof{76.49}{0.65}           & \resultnof{21.04}{1.08}           & \resultnof{71.54}{0.72}\\
\multirow{-4}{*}{\rotatebox[origin=c]{90}{2000}}& \hspace{0.2 cm }\textbf{$\hookrightarrow$SAM} & \resultnof{\textbf{33.76}}{1.21}  & \resultnof{\textbf{83.79}}{0.27}  & \resultnof{\textbf{22.53}}{1.56}  & \resultnof{\textbf{78.35}}{0.36}\\
\midrule
                                                & DualNet                                       & \resultnof{32.08}{1.55}           & \resultnof{80.26}{0.97}           & \resultnof{22.07}{2.08}            & \resultnof{74.53}{1.27}\\
\multirow{-2}{*}{\rotatebox[origin=c]{90}{5000}}& \hspace{0.2 cm }\textbf{$\hookrightarrow$SAM} & \resultnof{\textbf{36.44}}{0.77}  & \resultnof{\textbf{85.72}}{0.40}  & \resultnof{\textbf{24.83}}{2.01}   & \resultnof{\textbf{80.18}}{0.52}\\
\bottomrule
\end{tabular}
}
\label{tab:perf_competitor}
\end{table*}

\noindent \textbf{Effect of classification pre-training.}
Additionally, in order to demonstrate generalization capabilities of our attention-modulated strategy, and to ground our approach to the CL methods that exploit pre-training, we also compute performance when the classifier backbone and saliency encoder are pre-trained on a classification pre-text task (despite using classification-pretrained features appears to be in contrast to what it happens in the human brain). Differently from what describe in~\ref{sec:training}, here we use the same disjoint subset of ImageNet classes to train the backbone of the classifier, then we initialize the saliency encoder to the same weights.
Also in this setting, methods combined to SAM achieve better results, as shown in Table~\ref{tab:SaliencyPretrainClass}. However, the performance gain is lower than the one obtained with saliency pre-training. This is possibly due the fact that classification pre-trained features are better than saliency ones (as also evidenced by the general higher performance obtained with classification pre-training) and have reached their maximum capacity. 
These results confirm again the contribution of the \emph{forgetting-free} behaviour of the saliency prediction task to classification tasks. 

\begin{table*}[ht]
\caption{\textbf{Class-IL and Task-IL performance when the classifier backbone and saliency encoder are pre-trained on a classification task with classes different from those available in the continual learning settings.}}
\centering
\rowcolors{4}{gray!10}{white}
\normalsize{
\begin{tabular}{c|ccc|ccc} 
\toprule

\textbf{Model}                                  & \multicolumn{3}{c}{\textbf{Split Mini-ImageNet}}                                                    &\multicolumn{3}{c}{\textbf{Split FG-ImageNet}}\\ 
\textbf{Buffer}                                 & \textbf{1000}                     & \textbf{2000}               &\textbf{5000}                      &\textbf{1000}                      & \textbf{2000}                     & \textbf{5000}                     \\ 
\midrule
\arrayrulecolor{black}
                                                & \multicolumn{3}{c}{CLASS-IL}                                                                              & \multicolumn{3}{c}{CLASS-IL}                                                                              \\
\midrule
DER++                                           & \resultnof{30.35}{0.74}           & \resultnof{30.96}{0.59}     & \resultnof{32.55}{1.47}           & \resultnof{15.76}{0.58}           & \resultnof{16.61}{0.26}           & \resultnof{16.83}{0.44}           \\
\hspace{0.1 cm } {$\hookrightarrow$\textbf{SAM}}& \resultnof{ \textbf{31.20}}{2.39}        & \resultnof{ \textbf{33.91}}{2.31}  & \resultnof{ \textbf{37.91}}{1.07}  & \resultnof{ \textbf{17.06}}{1.51}  & \resultnof{ \textbf{20.43}}{2.11}  & \resultnof{ \textbf{22.53}}{0.82}  \\
\arrayrulecolor{gray}
\midrule
\arrayrulecolor{black}

ER-ACE                                          & \resultnof{42.33}{0.57}           & \resultnof{45.84}{0.50}     & \resultnof{48.77}{1.28}           & \resultnof{30.91}{1.02}           & \resultnof{34.09}{0.57}           & \resultnof{37.49}{0.47}           \\
\hspace{0.1 cm } {$\hookrightarrow$\textbf{SAM}}& \resultnof{\textbf{46.56}}{1.10}  & \resultnof{ \textbf{50.52}}{0.69}  & \resultnof{ \textbf{53.23}}{0.35}  & \resultnof{ \textbf{32.46}}{1.09}  & \resultnof{ \textbf{36.08}}{1.60}  & \resultnof{ \textbf{40.73}}{0.84}  \\
\midrule
                                                & \multicolumn{3}{c}{TASK-IL}                                                                               & \multicolumn{3}{c}{TASK-IL}                                                                               \\
\midrule
DER++                                           & \resultnof{ \textbf{89.98}}{0.75}        & \resultnof{ \textbf{91.14}}{0.20}  & \resultnof{ \textbf{91.37}}{0.10}  & \resultnof{ \textbf{83.87}}{0.81}  & \resultnof{ \textbf{85.61}}{0.29}  & \resultnof{ \textbf{86.19}}{0.21}  \\
\hspace{0.1 cm } {$\hookrightarrow$\textbf{SAM}}& \resultnof{89.34}{0.54}           & \resultnof{90.47}{0.32}     & \resultnof{91.36}{0.30}           & \resultnof{82.34}{0.54}           & \resultnof{84.04}{0.40}           & \resultnof{84.83}{0.32}           \\
\arrayrulecolor{gray}
\midrule
\arrayrulecolor{black}
ER-ACE                                          & \resultnof{88.28}{0.50}           & \resultnof{90.14}{0.05}     & \resultnof{91.23}{0.13}           & \resultnof{82.83}{0.40}           & \resultnof{ \textbf{85.39}}{0.38}  & \resultnof{ \textbf{87.29}}{0.08}  \\
\hspace{0.1 cm } {$\hookrightarrow$\textbf{SAM}}& \resultnof{ \textbf{89.99}}{0.46}        & \resultnof{\textbf{90.83}}{0.20}  & \resultnof{\textbf{91.84}}{0.08}  & \resultnof{\textbf{82.94}}{1.15}  & \resultnof{84.25}{0.95}           & \resultnof{86.51}{0.25}           \\
\bottomrule
\end{tabular}}
\label{tab:SaliencyPretrainClass}
\end{table*}

\begin{figure*}[ht!]
     \centering
        \includegraphics[width=0.32\linewidth]{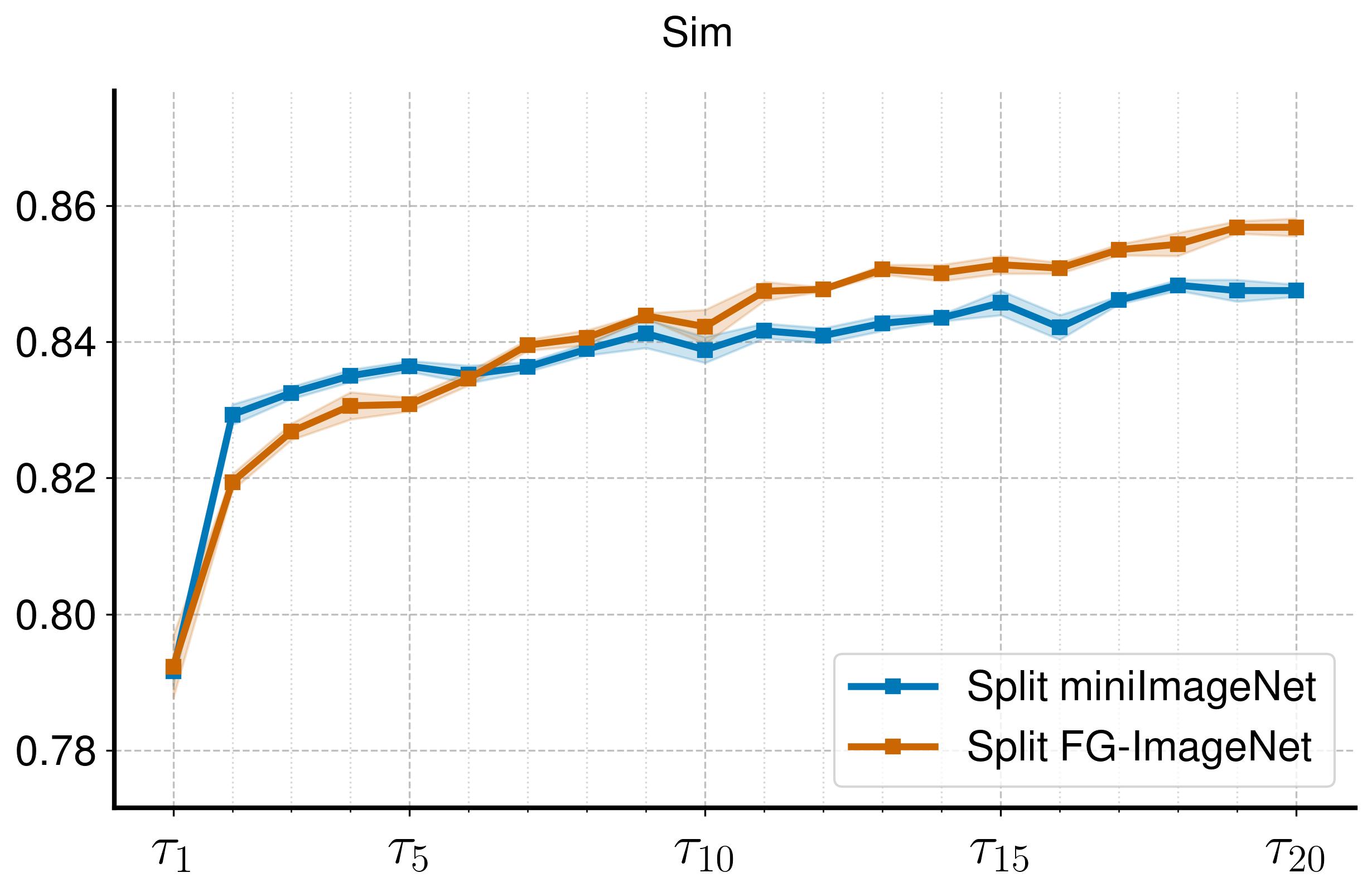}
         \includegraphics[width=0.32\linewidth]{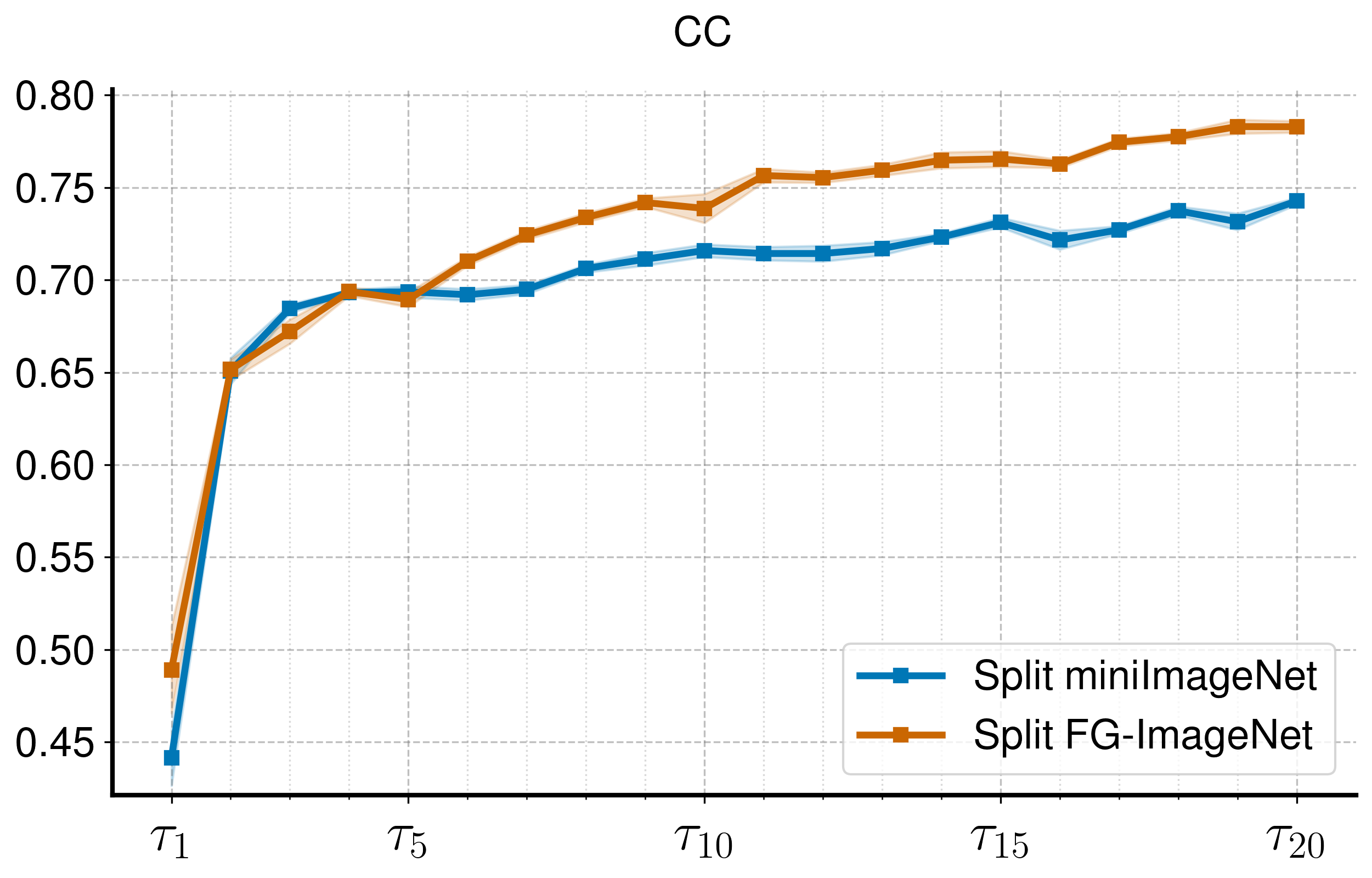}
         \includegraphics[width=0.32\linewidth]{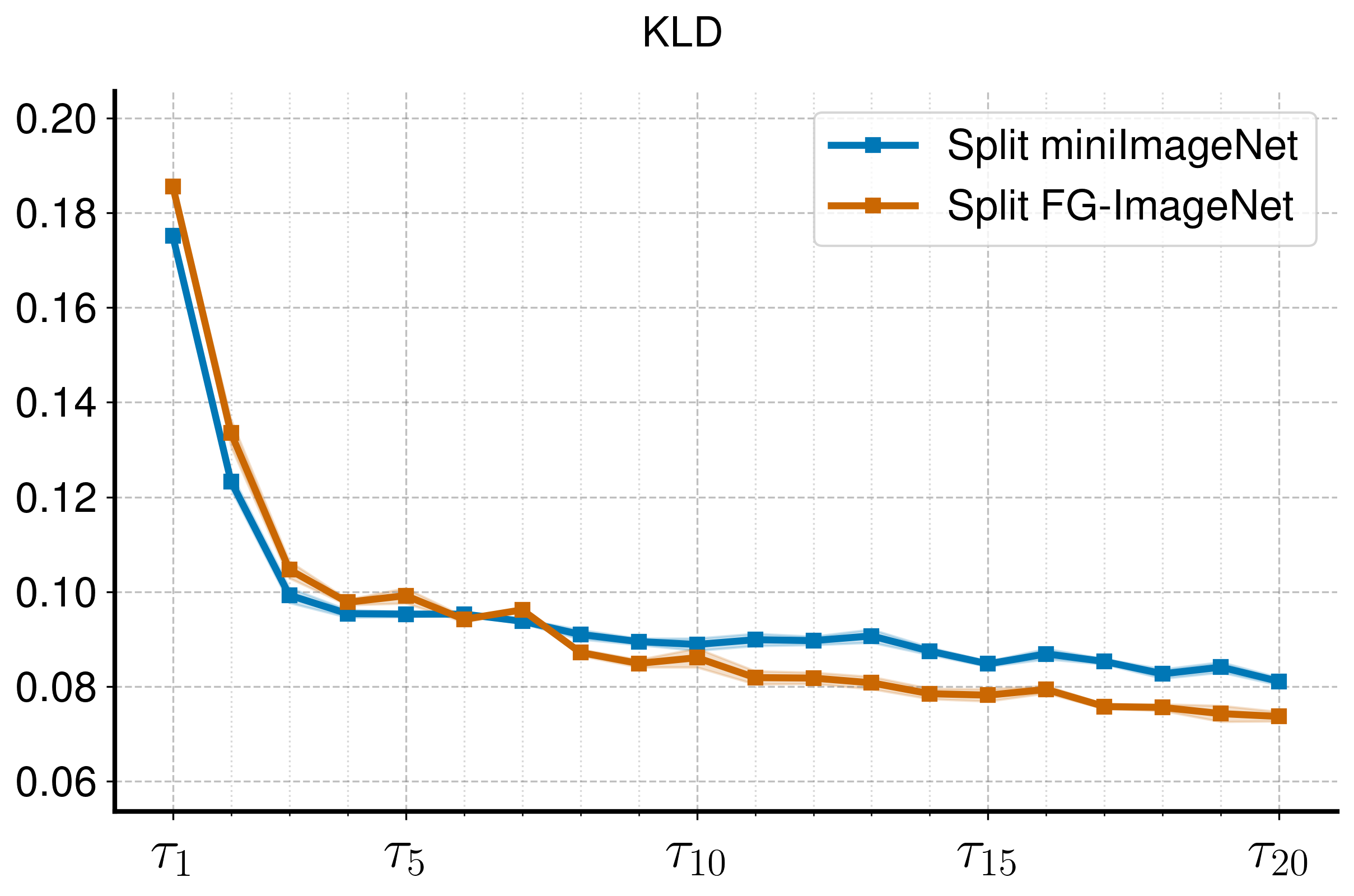}
        \caption{\textbf{Saliency prediction accuracy, measured in terms of Sim, CC and KLD metrics, in continual learning settings on the on Split miniImagenet and Split FG-Imagenet benchmarks. }}
        \label{fig:saliency_metrics}
\end{figure*}

\begin{figure*}
    \centering      
     \begin{subfigure}[b]{\textwidth}
         \centering
         \includegraphics[width=0.18\linewidth]{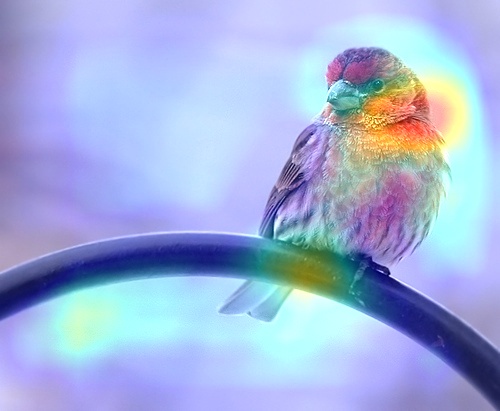}
         \includegraphics[width=0.18\linewidth]{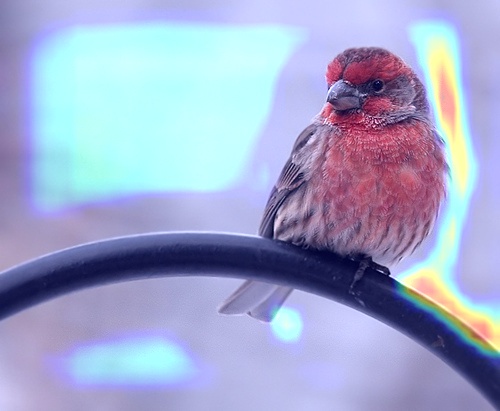}
         \includegraphics[width=0.18\linewidth]{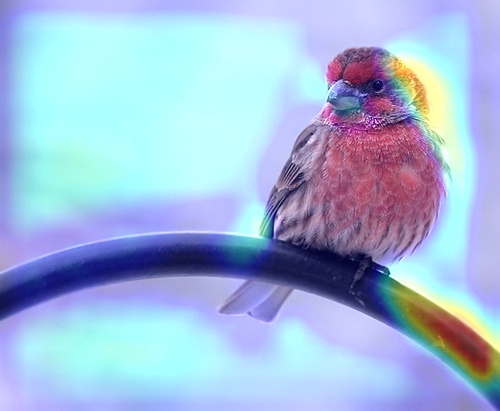}
         \includegraphics[width=0.18\linewidth]{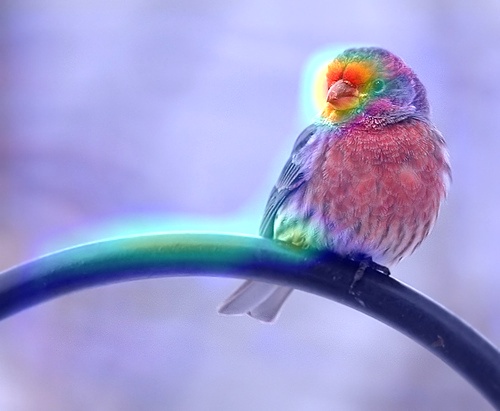}
         \includegraphics[width=0.18\linewidth]{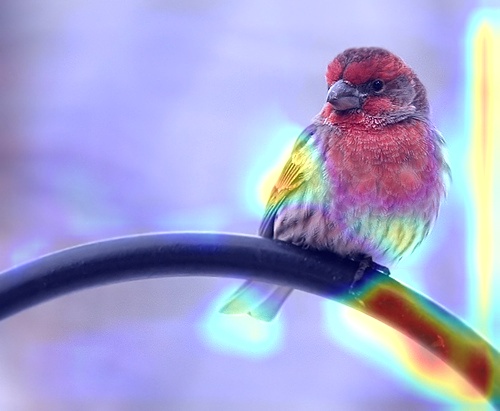}
     \end{subfigure}
     \\
     \vspace{0.1 cm}
     \begin{subfigure}[b]{\textwidth}
         \centering
         \includegraphics[width=0.18\linewidth]{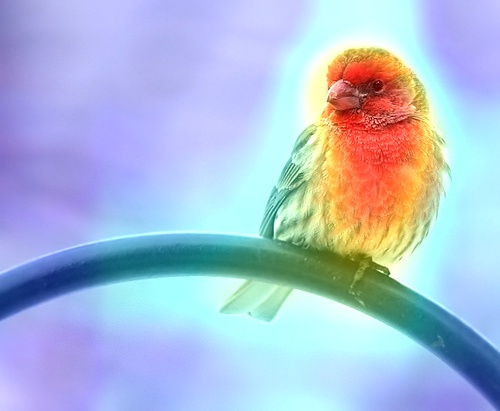}
         \includegraphics[width=0.18\linewidth]{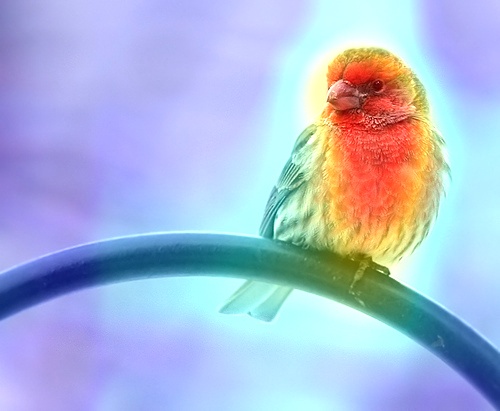}
         \includegraphics[width=0.18\linewidth]{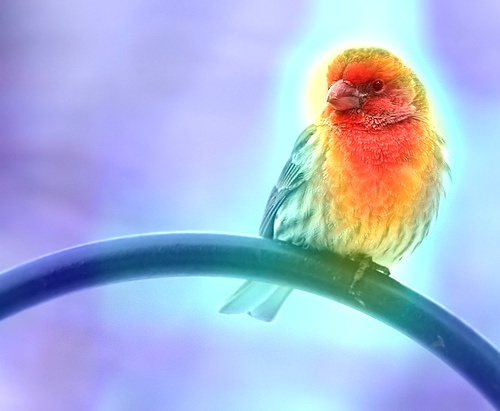}
         \includegraphics[width=0.18\linewidth]{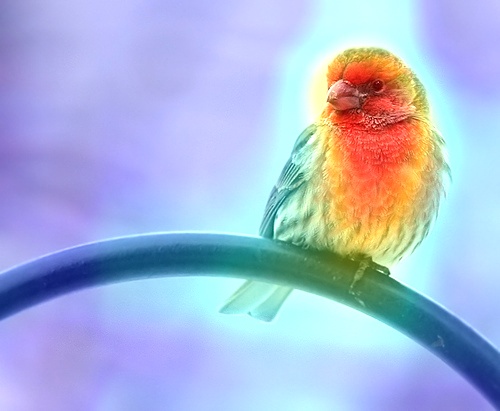}
         \includegraphics[width=0.18\linewidth]{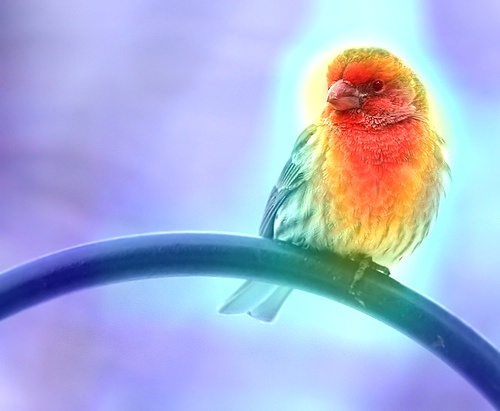}
     \end{subfigure}
     \makebox[0.18\linewidth]{$\tau_{1}$}
     \makebox[0.18\linewidth]{$\tau_{5}$}
     \makebox[0.18\linewidth]{$\tau_{10}$}
     \makebox[0.18\linewidth]{$\tau_{15}$}
     \makebox[0.18\linewidth]{$\tau_{20}$}
    \caption{\textbf{Qualitative comparison of attribution maps computed through GradCAM (first row) and the saliency maps produced by the saliency predictor $S$ (second row) during a continual training on a sequence of 20 tasks.} GradCAM attributions maps show significant forgetting, while saliency maps tend steadily to improve while training.}
    \label{fig:xai_vs_sal} \vspace{-0.5cm}
\end{figure*}

\noindent \textbf{Saliency Prediction Maps vs Attribution Maps.}
We also compute the saliency metrics obtained by our predictor $S$ in the considered class-incremental setting. In particular, we use three widely used metrics for image saliency prediction~\cite{bylinskii2018different}: Persons's Correlation Coefficient (CC), Similarity (Sim) and Kullback-Leibler divergence (KLD). As shown in Fig.~\ref{fig:saliency_metrics}, all metrics do not degrade as new tasks are processed, but rather they exhibit a trend of enhancement with the number of CL tasks. 
This behaviour is further corroborated by the qualitatively results shown Fig.~\ref{fig:xai_vs_sal} (second row). The predicted saliency maps shows no forgetting
when training on a sequence of twenty tasks (from $\tau_0$ to $\tau_{20}$). 
Conversely, the attribution maps computed through Grad-CAM~\cite{selvaraju2019grad} significantly deteriorates, showing a high level of forgetting.
These results thus demonstrate that pairing a saliency prediction with a classifier yields better results that storing attribution maps as done in~\cite{ebrahimi2021remembering,
selvaraju2019grad,
saha2023saliency}.

\noindent{\textbf{Cost analysis.}} We finally perform cost analysis to assess the efficiency of our SAM approach compared to existing methods that employ two branches, i.e., TwF~\cite{boschini2022transfer} and DualNet~\cite{pham2021dualnet}. 
It is important to note that in a continual learning settings, efficiency at training times might be more relevant than the one at inference times as the main assumption is of a deep model that keeps training from an infinite stream of data. The comparison is carried out on an NVIDIA A100 and using the ResNet18 backbone for all models. The results in Table~\ref{tab:cost_analysis} reveals that SAM is much more efficient than DualNet and TwF at training time, while it shows higher costs at inference time (but also an accuracy gain of $\sim$10 points).

\begin{table}[h!]
\centering
\caption{\textbf{Efficiency analysis}. Comparison of training and inference times and parameters between SAM, DualNet and TwF.}
\rowcolors{2}{white}{gray!10}
\normalsize{
\begin{tabular}{l|rrr} 
\toprule
\textbf{Metric} & \textbf{DualNet~\cite{pham2021dualnet}} & \textbf{TwF~\cite{boschini2022transfer}} & \textbf{SAM} \\
\midrule
Train params & 16 M & 58 M & 23 M \\
Train time & $\sim$ 6.5 h & $\sim$ 3.0 h & $\sim$ 1.0 h \\
Inference params & 16 M & 11 M & 22 M \\
Inference time & 3.45 ms & 3.15 ms & 7.50 ms \\
\bottomrule
\end{tabular}}
\label{tab:cost_analysis}
\end{table}

\subsection{Ablation Study}
\begin{figure*}
     \centering
     \begin{subfigure}[b]{0.49\textwidth}
         \centering
         \caption{Split Mini-ImageNet}
         \includegraphics[width=0.45\linewidth]{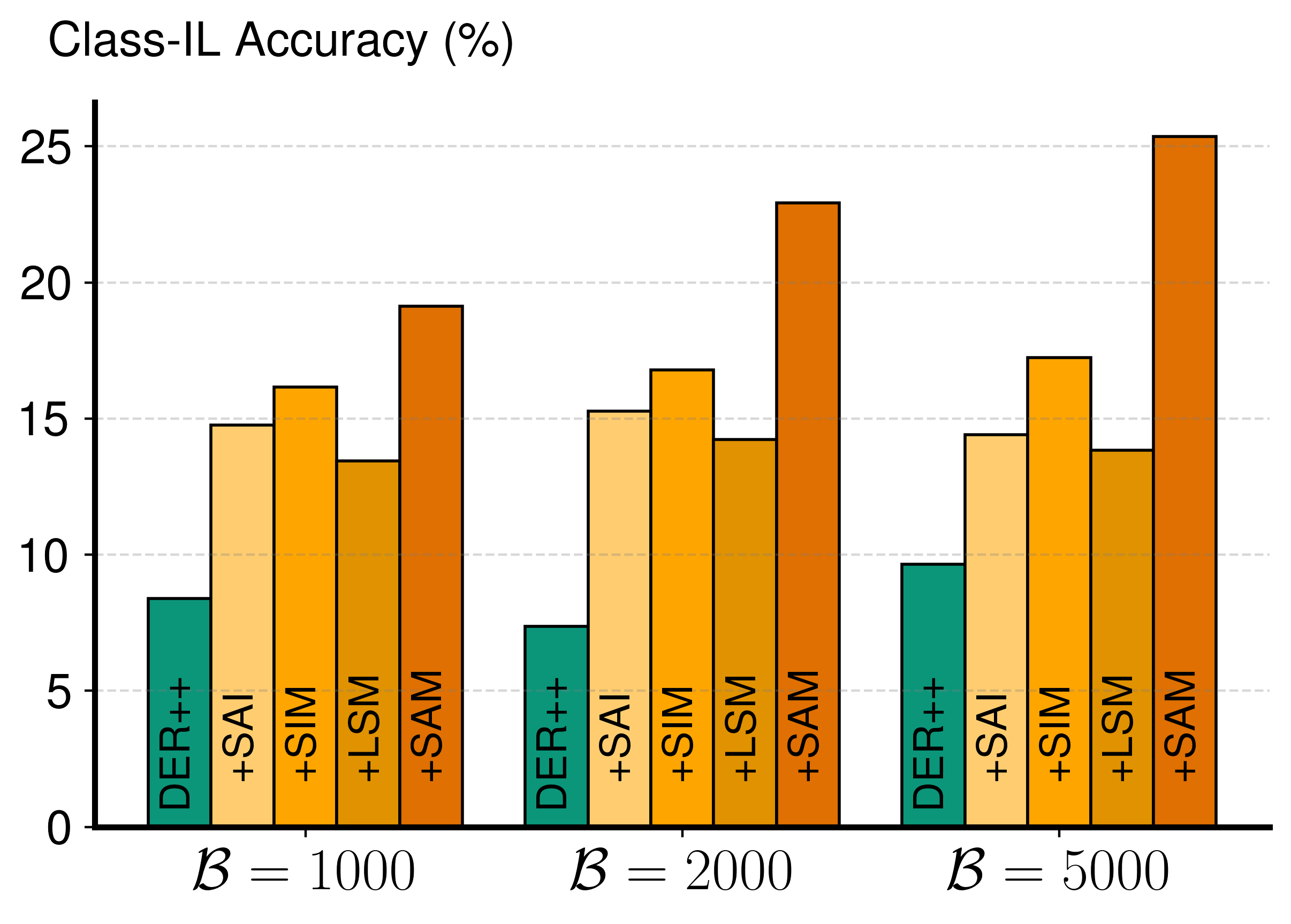}
         \includegraphics[width=0.45\linewidth]{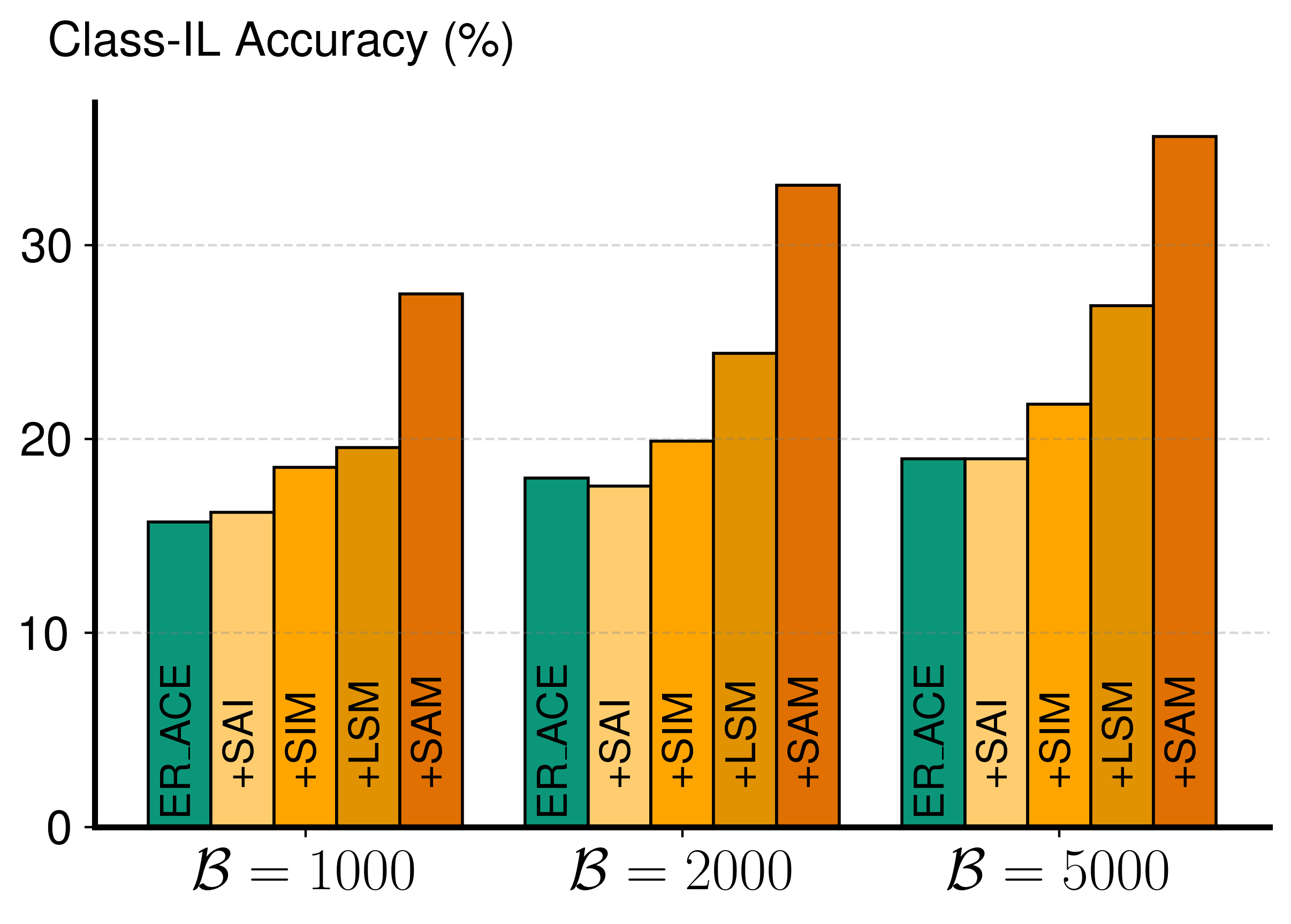}
     \end{subfigure}
     \begin{subfigure}[b]{0.49\textwidth}
         \centering
         \caption{Split FG-ImageNet}
         \includegraphics[width=0.45\linewidth]{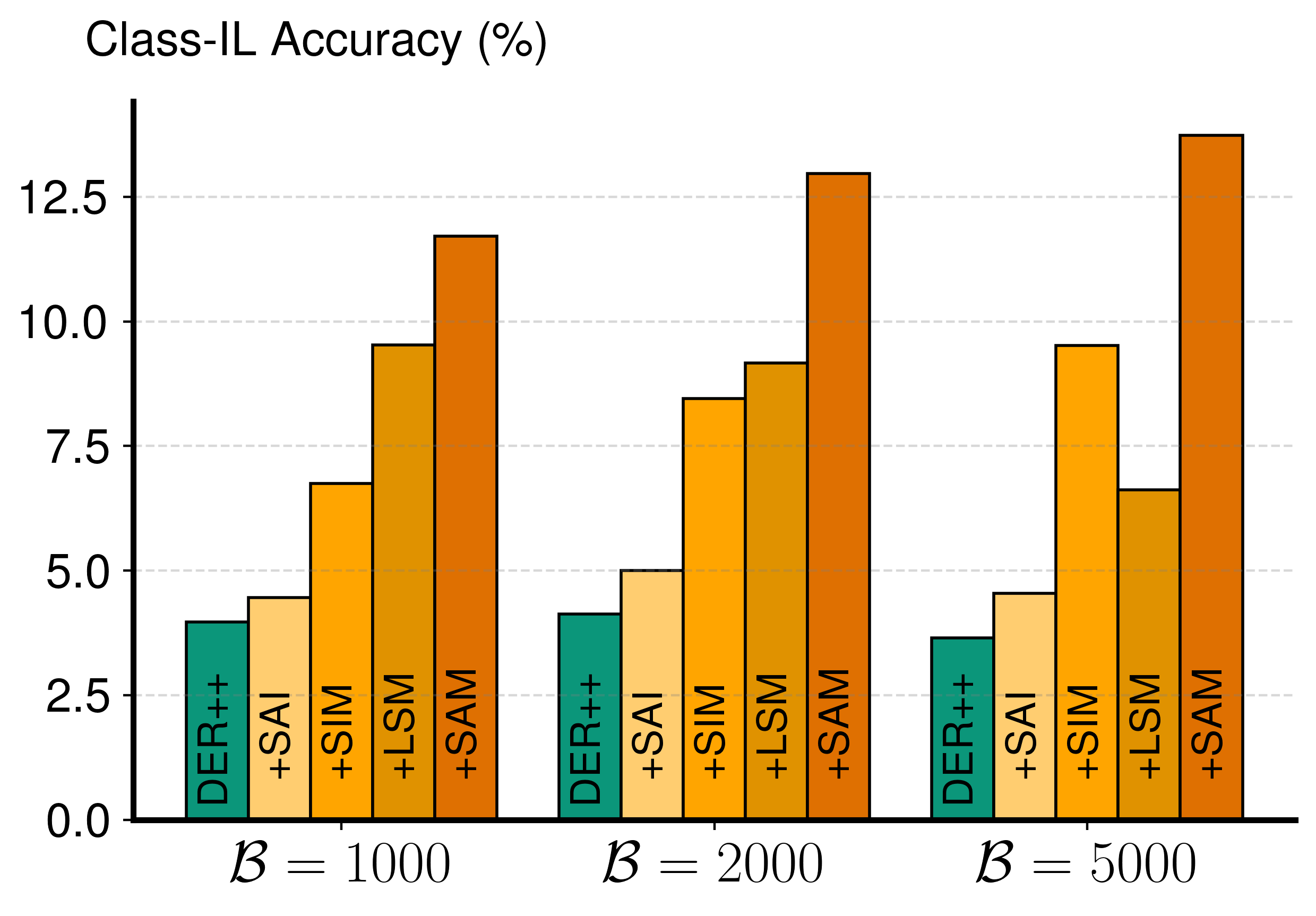}
         \includegraphics[width=0.45\linewidth]{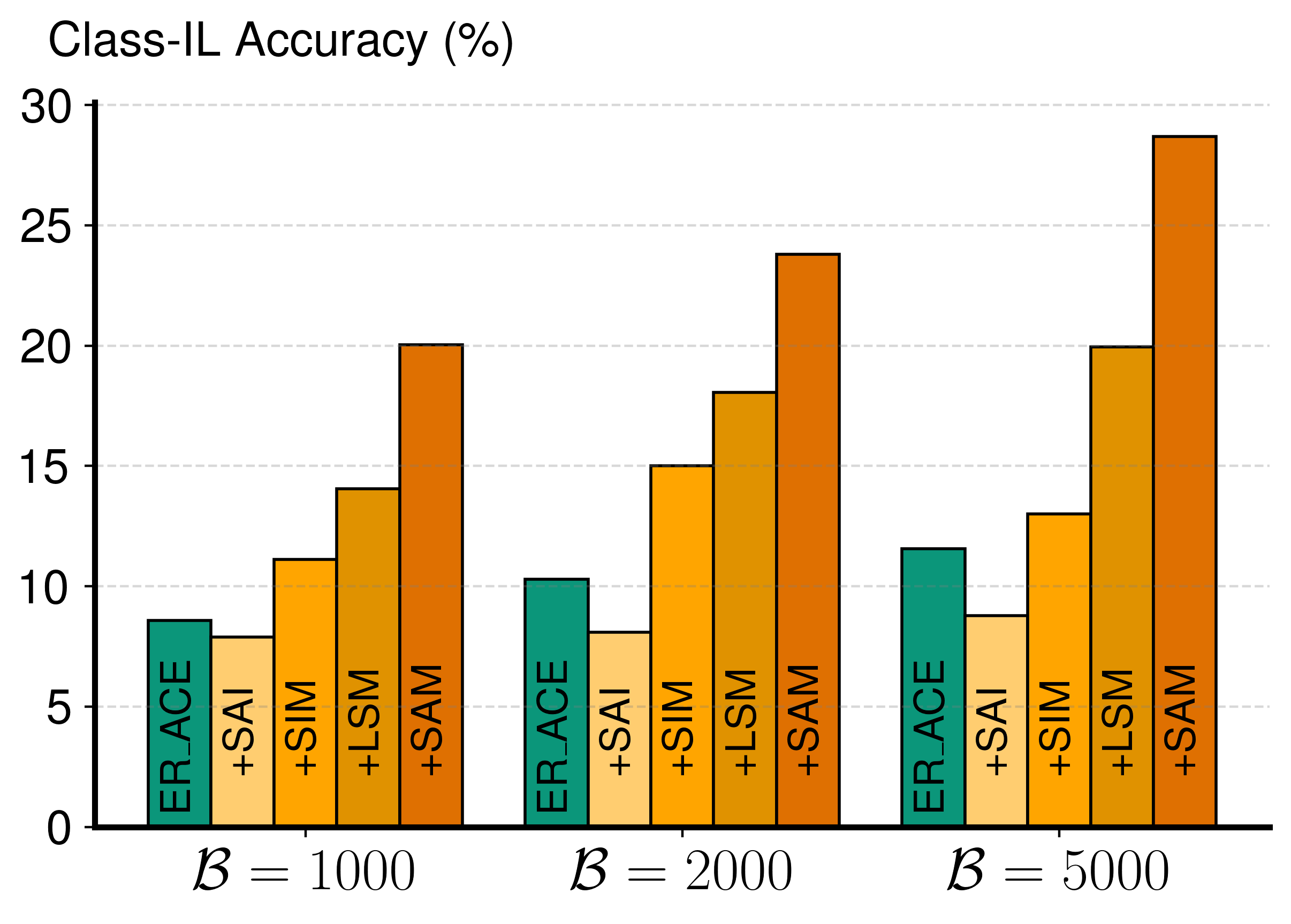}
     \end{subfigure}
     \\
     \begin{subfigure}[b]{0.49\textwidth}
         \centering
         \includegraphics[width=0.45\linewidth]{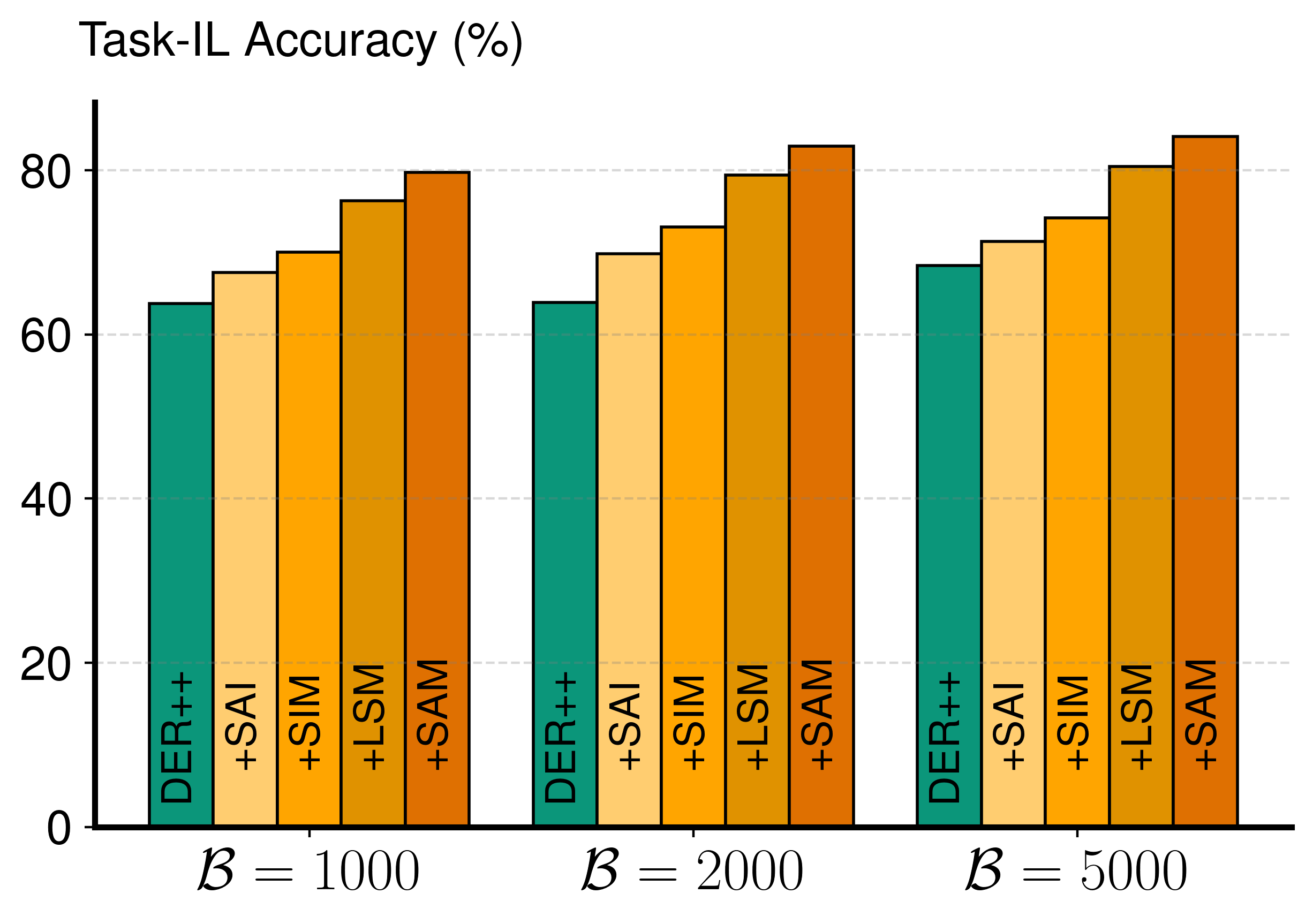}
         \includegraphics[width=0.45\linewidth]{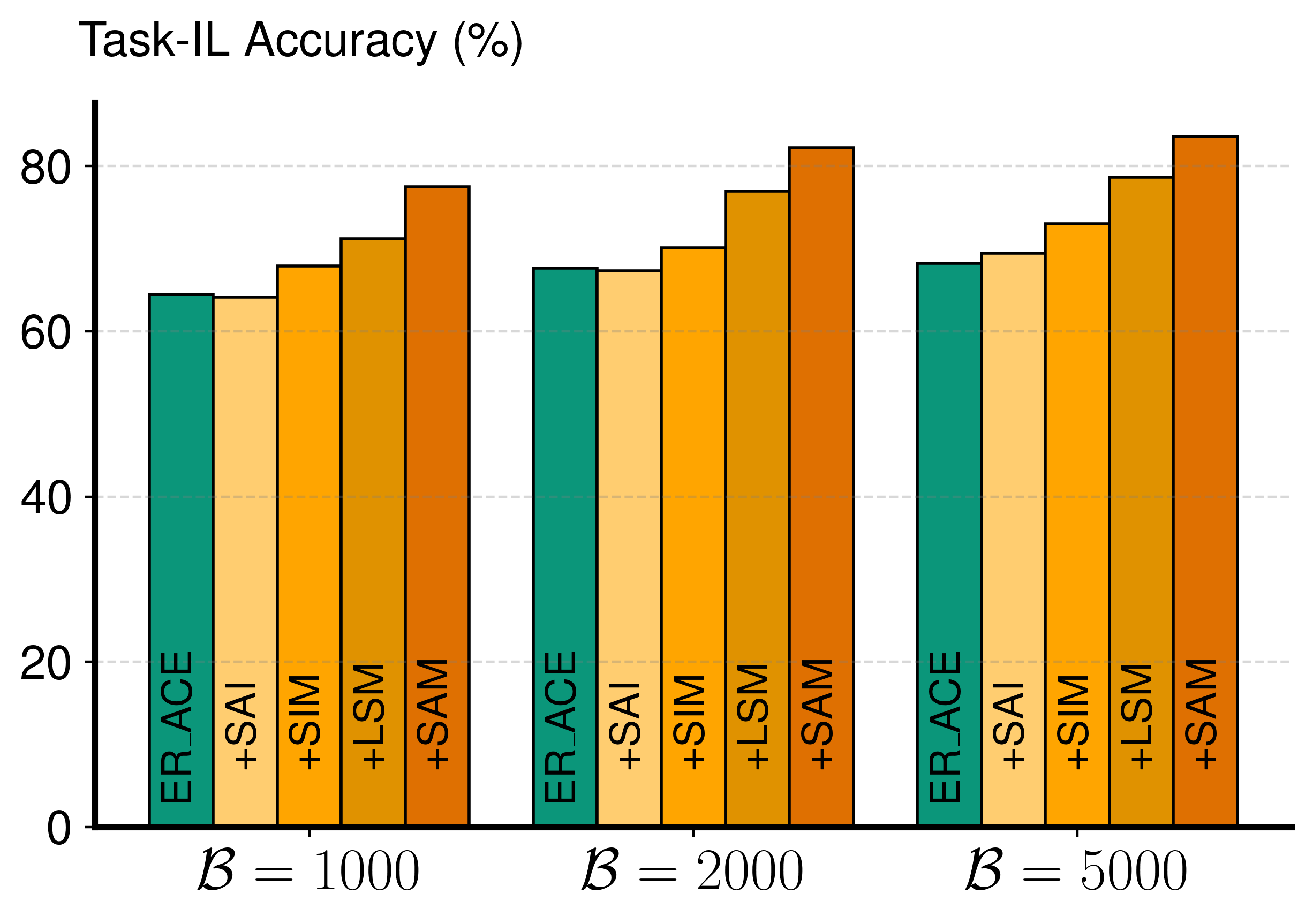}
     \end{subfigure}
     \begin{subfigure}[b]{0.49\textwidth}
         \centering
         \includegraphics[width=0.45\linewidth]{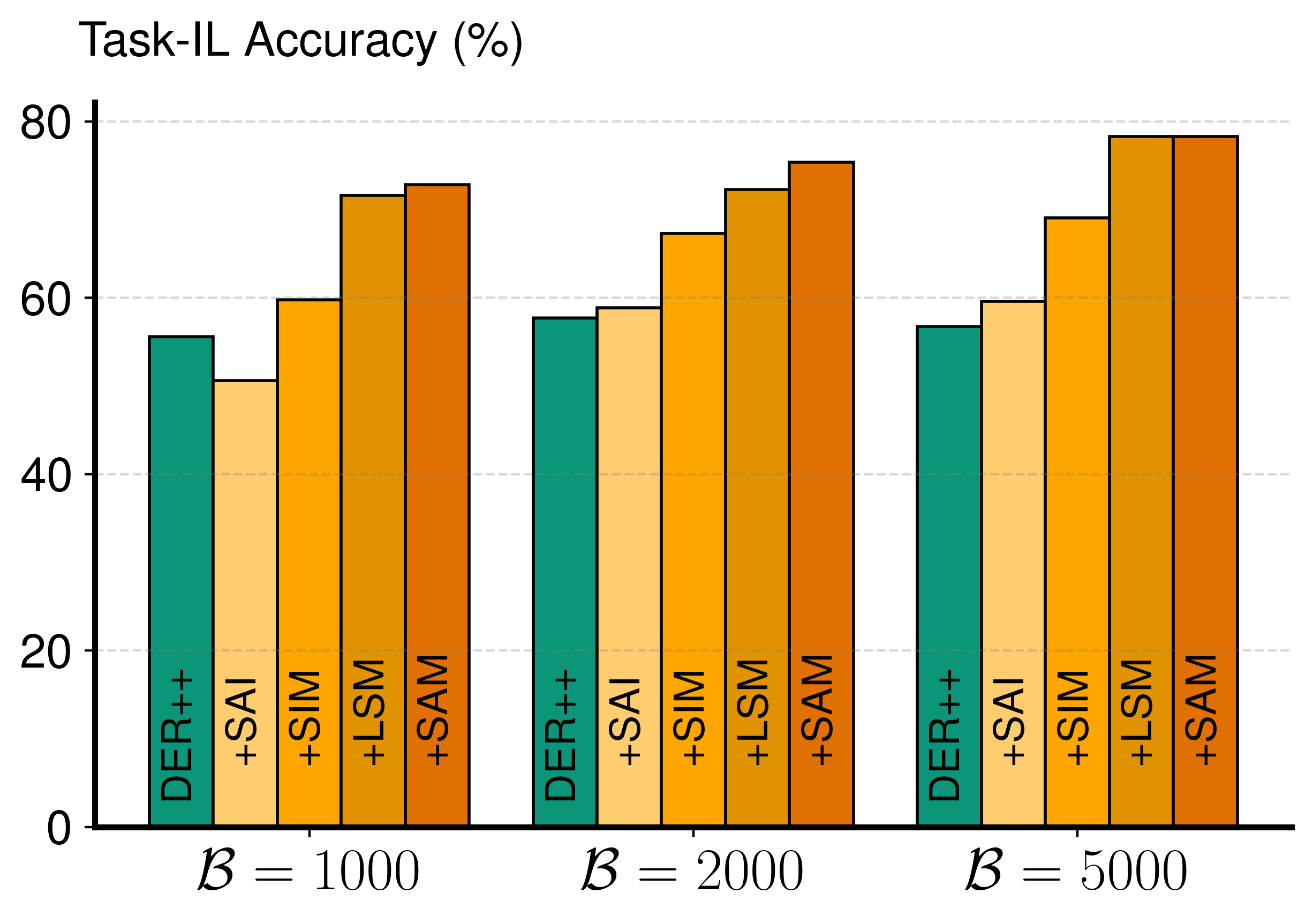}
         \includegraphics[width=0.45\linewidth]{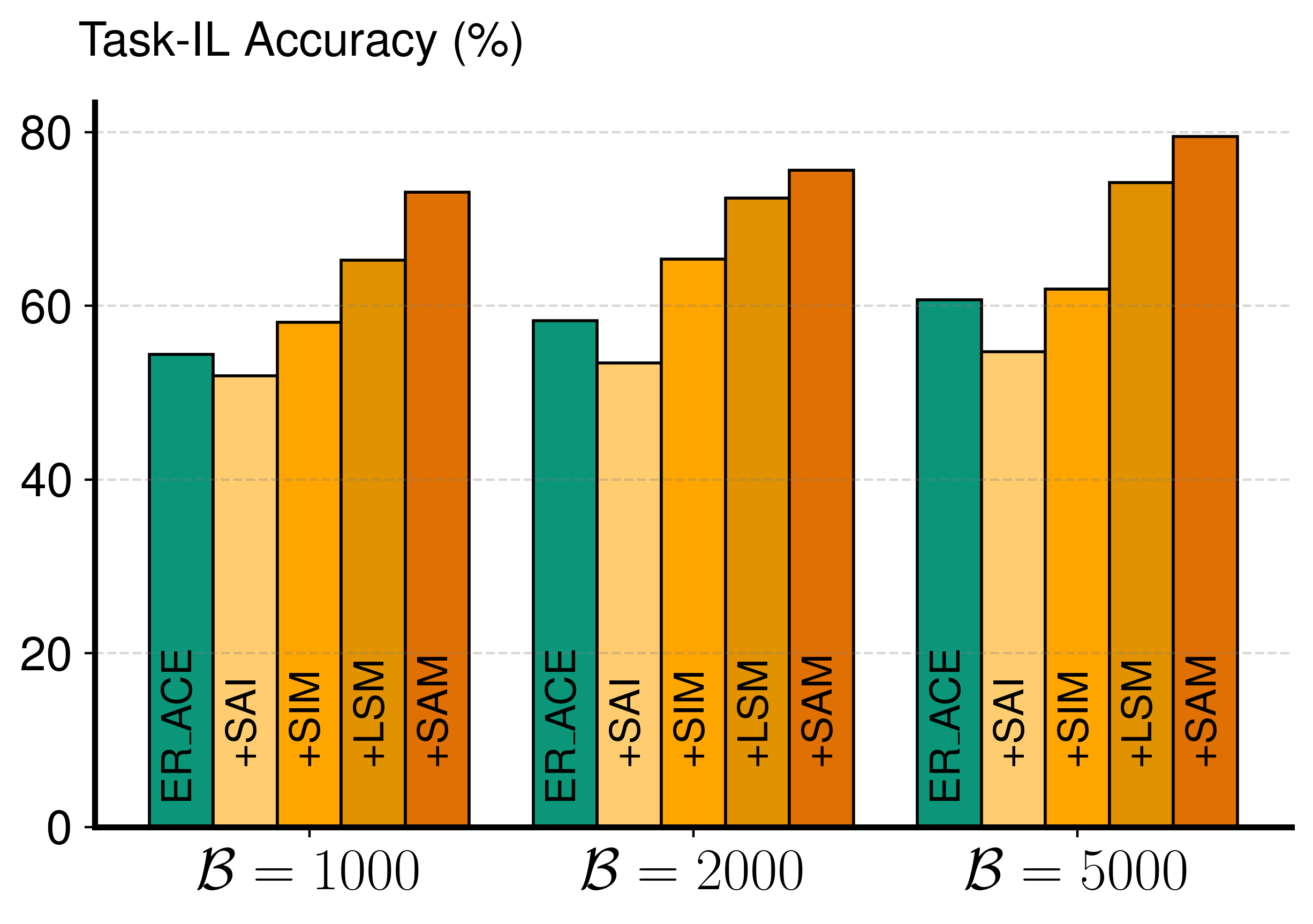}
     \end{subfigure}
        \caption{ \textbf{Comparison of SAM to alternative saliency integration strategies}. \textbf{SIM} modulates input images by saliency maps. \textbf{SAI} provides saliency maps as an additional input channel to the classification network. \textbf{LSM} merges classification and saliency features through a learnable convolutional layer.}
        \label{fig:saliency}
\end{figure*}
The proposed strategy is grounded on cognitive neuroscience literature, according to which selective attention modulates neuronal responses of all layers involved in the categorization process, in a multiplicative fashion. Our next experiments are meant to assess whether this hypothesis (i.e., feature modulation through multiplication for all classification layers) is optimal also for artificial neural networks, or if other integration modalities of saliency information may be equally effective.  
We thus compare our SAM strategy with the following baselines, all exploiting saliency information in different ways:
\begin{itemize}[noitemsep,nolistsep,leftmargin=*]
    \item \textbf{Saliency-based input modulation (SIM)}: the input image is multiplied by the corresponding estimated saliency map (thus highlighting salient regions only).
    \item \textbf{Saliency as additional input (SAI)}: we modify the classification network to receive as input a 4D data tensor, with the saliency map concatenated to RGB channels. 
    \item \textbf{Learning saliency-based modulation (LSM)}: rather than multiplying classification features $\zz_{i-1}^{(c)}$ and saliency features $\zz_{i-1}^{(s)}$ (see Eq.~\ref{eq:sam}), we feed them to convolutional layer with 1$\times$1 kernel to produce $\zz_{i}^{(c)}$, and let the model learn the corresponding parameters. 
\end{itemize}
\noindent Fig.~\ref{fig:saliency} reports the results of this analysis, using DER++ and ER-ACE as baseline methods, and clearly indicates the superiority the SAM strategy to other saliency integration variants. However, it is interesting to note that saliency helps classification performance in all cases, demonstrating its usefulness for continual learning tasks. We argue that this is due to the intrinsic nature of saliency prediction, which we found to be i.i.d. with respect to the data stream. 

We then investigate whether the impact of attention-driven modulation is uniform across the backbone layers. To this aim, we define a positional binary coding scheme, controlling the application of the SAM strategy at the predefined points of the network (see Sect.~\ref{sec:training}): if position $i$ of the coding scheme is 1, then the $i$-th feature modulation point is enabled, i.e., features from the $i$-th block of the classification network are multiplied by the features of the $i$-th block  of the saliency network. 
Results are reported in Table~\ref{tab:abl_coding} for both DER++ and ER-ACE, and indicate that the best strategy is to modulate the features of all classification layers through the corresponding saliency ones, similarly to what neurophysiological evidence reports~\cite{pmid10376597,pmid15120065}.

\begin{table*}[ht]
\centering
\caption{\textbf{Performance comparison when applying SAM to DER++ and ER-ACE} at different layers of the ResNet-18 backbone, with buffer size 2000.}
\rowcolors{3}{white}{gray!10}
\renewcommand{\arraystretch}{1.2}
\normalsize{
\begin{tabular}{c|cc|cc}
\toprule
 \textbf{SAM}     & \multicolumn{2}{c}{\textbf{Split Mini-ImageNet}}                              &\multicolumn{2}{c}{\textbf{Split FG-ImageNet}}\\
\textbf{Scheme}                    & Class-IL                           & Task-IL                           & Class-IL                          & Task-IL\\
\midrule
\arrayrulecolor{black}
\multicolumn{5}{c}{\textit{DER++}} \\
\midrule
\textbf{1 1 1 0 0}                 &\resultnof{12.97}{2.62}             & \resultnof{74.55}{3.62}           & \resultnof{6.54}{0.67}            & \resultnof{67.34}{1.38}\\
\textbf{1 1 1 1 0}                 &\resultnof{17.46}{1.02}             & \resultnof{80.15}{0.34}           & \resultnof{8.77}{1.45}            & \resultnof{71.51}{2.92}\\
\textbf{1 1 1 1 1}                 &\resultnof{\textbf{22.92}}{2.25}    & \resultnof{\textbf{82.97}}{0.25}  & \resultnof{\textbf{12.97}}{1.62}  & \resultnof{\textbf{75.40}}{2.29}\\
\midrule
\multicolumn{5}{c}{\textit{ER-ACE}} \\
\midrule
\textbf{1 1 1 0 0}                 & \resultnof{23.72}{0.77}            & \resultnof{74.15}{1.38}           & \resultnof{18.08}{0.96}           & \resultnof{70.44}{2.08}\\
\textbf{1 1 1 1 0}                 & \resultnof{26.44}{2.33}            & \resultnof{77.14}{2.73}           & \resultnof{16.55}{2.55}           & \resultnof{67.32}{5.07}\\
\textbf{1 1 1 1 1}                 & \resultnof{\textbf{33.09}}{1.28}   & \resultnof{\textbf{82.22}}{0.96}  & \resultnof{\textbf{23.80}}{2.11}  & \resultnof{\textbf{75.60}}{2.28}\\
\bottomrule
\end{tabular}
}
\label{tab:abl_coding}
\end{table*}

\subsection{Model Robustness}
We finally assess the robustness of the SAM strategy in dealing with \emph{spurious features} and \emph{adversarial attacks}.

Spurious features are information that correlates well with labels in training data but not in test data (e.g., in a classification task between birds and dogs, training with yellow birds and black dogs only), leading to low generalization~\cite{spurious}. This effect is exacerbated in continual learning settings, where the covariate shift between train data and test data increases as new tasks come in. Thus, we measure to what extent our SAM strategy can mitigate the tendency of learning methods to exploit spurious features to solve classification tasks.
We crafted an ad-hoc benchmark consisting of ten classes from ImageNet. For each class, we added a class signature for training images, leaving the test images unaltered. In detail, we modified each training image by increasing the brightness of all pixels by a class-dependent offset, computed as $5(c+1)$ (in a 0-255 brightness range), where $c$ is a numeric class label. We then define five continual learning tasks with two classes each.
We then compare ER-ACE to the corresponding SAM-enabled variant and ground its performance with the one obtained when it is trained with original images (i.e., without enforcing spurious features in the data). Results in Table~\ref{tab:perf_robustness} show that SAM effectively limits the possibility for the classifier to use spurious features, resulting in a more robust and generalizing model. The drop of performance (about 22 percent points) observed between training with the original data and training with data biased by spurious features is almost completely recovered when SAM is used. 
\begin{table}[t]
\centering
\caption{\textbf{Effect of the SAM strategy in the presence of spurious features}. The $\mathcal{SF}$ apex indicates that the method is trained on the biased dataset containing spurious features, while the one without apex when ER-ACE it trained on the original, spurious-free, dataset.}
\rowcolors{2}{gray!10}{white}
\renewcommand{\arraystretch}{1.2}
\normalsize{
\begin{tabular}{l|cc} 
\toprule
\textbf{Method}    & Class-IL & Task-IL \\
\midrule
\arrayrulecolor{black}
ER-ACE                                          & \resultnof{50.07}{3.88}               & \resultnof{86.77}{1.63} \\ 
ER-ACE$^{\mathcal{SF}}$                         & \resultnof{28.46}{3.46}               & \resultnof{74.40}{4.37} \\ 
\hspace{0.2 cm }\textbf{$\hookrightarrow$SAM}   & \resultnof{\textbf{44.08}}{3.67}      & \resultnof{\textbf{83.04}}{3.06}  \\
\bottomrule
\end{tabular}}
\label{tab:perf_robustness}
\end{table}

\begin{figure}[ht]
\centering
    \includegraphics[width=0.6\textwidth]{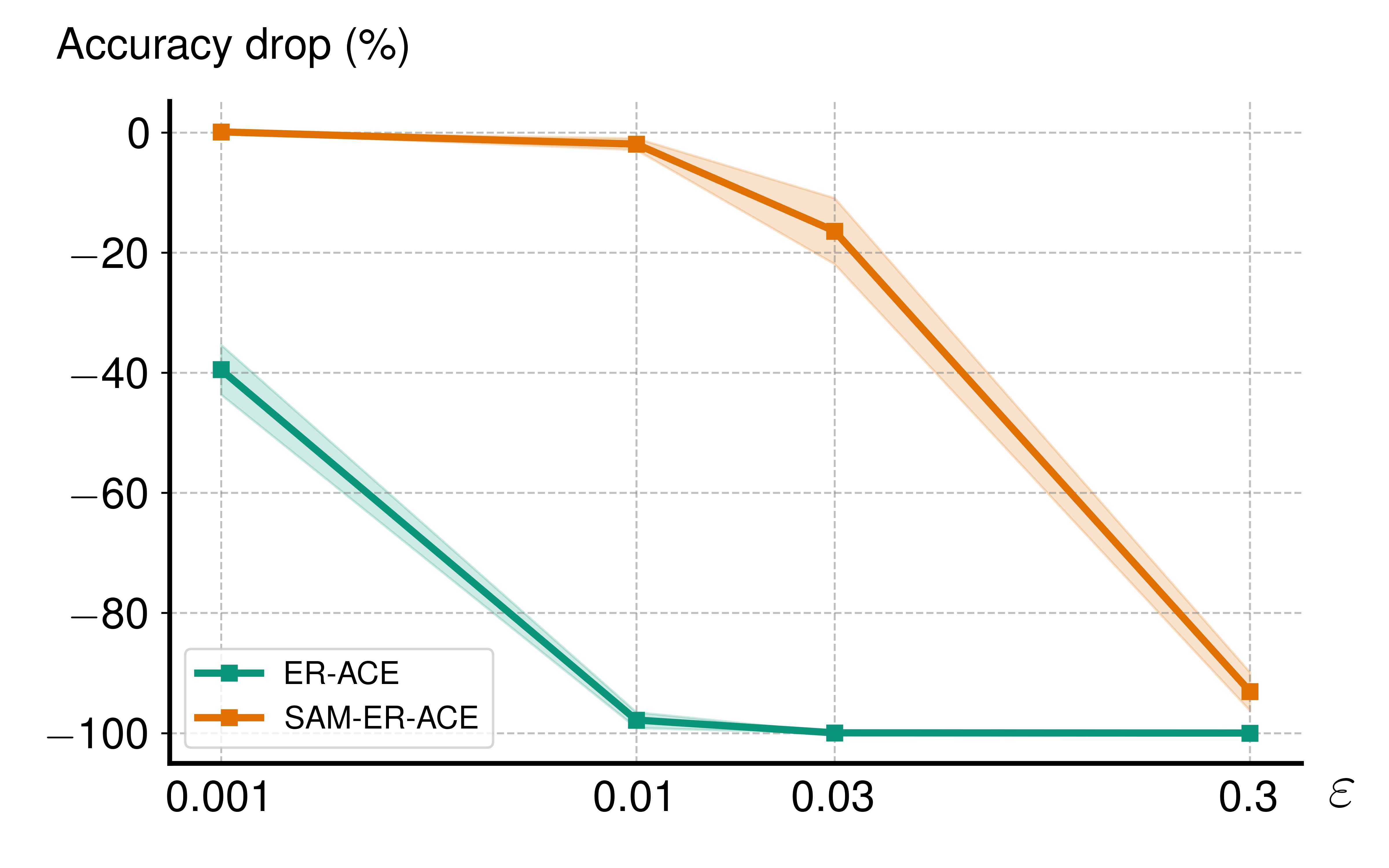}
    \caption{\textbf{Robustness to adversarial attacks}. ER-ACE baseline drops even with small attacks, while SAM significantly enhances robustness.}
    \label{fig:atck}
\end{figure}

Finally, we evaluate the robustness of SAM against adversarial perturbations of the input space. To this aim, we apply the Projected Gradient Descent (PGD) attack~\cite{madry2018towards} with different $\varepsilon$ values (determining the strength of the attack) and compare the average performance drop experienced by ER-ACE, in its original version and when combined with SAM. We conduct the evaluation on both Split Mini-ImageNet and Split FG-ImageNet, repeating each experiment three times. As shown in Figure~\ref{fig:atck}, SAM considerably improves model stability, counteracting perturbations by regularizing classification features with saliency ones.

\section{Conclusion}
We presented SAM, a biologically-inspired selective attention-driven modulation strategy for online continual learning, which regularizes classification features using visual saliency, effectively reducing forgetting. The proposed approach, grounded on neurophysiological evidence, significantly improves performance of state-of-the-art OCL methods, and has been shown to be superior to other multi-branch solutions, either biologically-inspired (e.g., DualNet) or based on feature attention mechanisms (e.g., TwF).

Our results confirm that adapting neurophysiological processes into current machine learning techniques is a promising direction to bridge the gap between humans and machines. Future research directions will address both limitations and extensions of the proposed approach. Indeed, while SAM is model-agnostic, its formulation requires that the saliency encoder and the classifier are architecturally identical. The application to heterogeneous networks will be explored by defining or learning a mapping between activations at different network stages. Moreover, our finding that saliency prediction is i.i.d.\ with respect to classification distribution shifts will lead to investigate whether other low-level visual tasks enjoy this property.

\section*{Acknowledgments}
This research was supported by Ministero dell'Università e della Ricerca (MUR) - PRIN 2020, project: ``LEGO.AI: LEarning the Geometry of knOwledge in AI systems'', n. 2020TA3K9N, CUP: E63C20011250001, and 
Matteo Pennisi is a PhD student enrolled in the National PhD in Artificial Intelligence, XXXVII cycle, course on Health and life sciences, organized by Università Campus Bio-Medico di Roma.



\begin{thebibliography}{10}
\providecommand{\url}[1]{#1}
\csname url@samestyle\endcsname
\providecommand{\newblock}{\relax}
\providecommand{\bibinfo}[2]{#2}
\providecommand{\BIBentrySTDinterwordspacing}{\spaceskip=0pt\relax}
\providecommand{\BIBentryALTinterwordstretchfactor}{4}
\providecommand{\BIBentryALTinterwordspacing}{\spaceskip=\fontdimen2\font plus
\BIBentryALTinterwordstretchfactor\fontdimen3\font minus \fontdimen4\font\relax}
\providecommand{\BIBforeignlanguage}[2]{{%
\expandafter\ifx\csname l@#1\endcsname\relax
\typeout{** WARNING: IEEEtran.bst: No hyphenation pattern has been}%
\typeout{** loaded for the language `#1'. Using the pattern for}%
\typeout{** the default language instead.}%
\else
\language=\csname l@#1\endcsname
\fi
#2}}
\providecommand{\BIBdecl}{\relax}
\BIBdecl

\bibitem{bylinskii2018different}
Z.~Bylinskii, T.~Judd, A.~Oliva, A.~Torralba, and F.~Durand, ``What do different evaluation metrics tell us about saliency models?'' \emph{IEEE transactions on pattern analysis and machine intelligence}, 2018.

\bibitem{pmid7624455}
J.~L. McClelland, B.~L. McNaughton, and R.~C. O'Reilly, ``{{W}hy there are complementary learning systems in the hippocampus and neocortex: insights from the successes and failures of connectionist models of learning and memory},'' \emph{Psychol Rev}, vol. 102, no.~3, pp. 419--457, Jul 1995.

\bibitem{pmid27315762}
D.~Kumaran, D.~Hassabis, and J.~L. McClelland, ``{{W}hat {L}earning {S}ystems do {I}ntelligent {A}gents {N}eed? {C}omplementary {L}earning {S}ystems {T}heory {U}pdated},'' \emph{Trends Cogn Sci}, vol.~20, no.~7, pp. 512--534, Jul 2016.

\bibitem{pmid28292907}
J.~Kirkpatrick, R.~Pascanu, N.~Rabinowitz, J.~Veness, G.~Desjardins, A.~A. Rusu, K.~Milan, J.~Quan, T.~Ramalho, A.~Grabska-Barwinska, D.~Hassabis, C.~Clopath, D.~Kumaran, and R.~Hadsell, ``{{O}vercoming catastrophic forgetting in neural networks},'' \emph{Proc Natl Acad Sci U S A}, vol. 114, no.~13, pp. 3521--3526, Mar 2017.

\bibitem{10.5555/3295222.3295393}
D.~Lopez-Paz and M.~Ranzato, ``Gradient episodic memory for continual learning,'' in \emph{Proceedings of the 31st International Conference on Neural Information Processing Systems}, ser. NIPS'17.\hskip 1em plus 0.5em minus 0.4em\relax Red Hook, NY, USA: Curran Associates Inc., 2017, p. 6470–6479.

\bibitem{kemker2017fearnet}
R.~Kemker and C.~Kanan, ``Fearnet: Brain-inspired model for incremental learning,'' \emph{arXiv preprint arXiv:1711.10563}, 2017.

\bibitem{pham2021dualnet}
Q.~Pham, C.~Liu, and S.~Hoi, ``Dualnet: Continual learning, fast and slow,'' \emph{Advances in Neural Information Processing Systems}, 2021.

\bibitem{pmid22325196}
J.~J. DiCarlo, D.~Zoccolan, and N.~C. Rust, ``{{H}ow does the brain solve visual object recognition?}'' \emph{Neuron}, 2012.

\bibitem{pmid17344377}
A.~Kohn, ``{{V}isual adaptation: physiology, mechanisms, and functional benefits},'' \emph{J Neurophysiol}, 2007.

\bibitem{ramasesh2021anatomy}
V.~V. Ramasesh, E.~Dyer, and M.~Raghu, ``Anatomy of catastrophic forgetting: Hidden representations and task semantics,'' in \emph{International Conference on Learning Representations Workshop}, 2021.

\bibitem{pnas.0703913104}
\BIBentryALTinterwordspacing
J.~New, L.~Cosmides, and J.~Tooby, ``Category-specific attention for animals reflects ancestral priorities, not expertise,'' \emph{Proceedings of the National Academy of Sciences}, vol. 104, no.~42, pp. 16\,598--16\,603, 2007. [Online]. Available: \url{https://www.pnas.org/doi/abs/10.1073/pnas.0703913104}
\BIBentrySTDinterwordspacing

\bibitem{Borji2019}
\BIBentryALTinterwordspacing
A.~Borji, ``Saliency prediction in the deep learning era: Successes, limitations, and future challenges,'' 2018. [Online]. Available: \url{https://arxiv.org/abs/1810.03716}
\BIBentrySTDinterwordspacing

\bibitem{pmid10376597}
S.~Treue and J.~C. nez Trujillo, ``{{F}eature-based attention influences motion processing gain in macaque visual cortex},'' \emph{Nature}, vol. 399, no. 6736, pp. 575--579, Jun 1999.

\bibitem{pmid15120065}
J.~C. Martinez-Trujillo and S.~Treue, ``{{F}eature-based attention increases the selectivity of population responses in primate visual cortex},'' \emph{Curr Biol}, vol.~14, no.~9, pp. 744--751, May 2004.

\bibitem{linardos2021deepgaze}
A.~Linardos, M.~K{\"u}mmerer, O.~Press, and M.~Bethge, ``Deepgaze iie: Calibrated prediction in and out-of-domain for state-of-the-art saliency modeling,'' in \emph{Proceedings of the IEEE/CVF International Conference on Computer Vision}, 2021, pp. 12\,919--12\,928.

\bibitem{jiang2015salicon}
M.~Jiang, S.~Huang, J.~Duan, and Q.~Zhao, ``Salicon: Saliency in context,'' in \emph{Proceedings of the IEEE conference on computer vision and pattern recognition}, 2015, pp. 1072--1080.

\bibitem{droste2020unified}
R.~Droste, J.~Jiao, and J.~A. Noble, ``Unified image and video saliency modeling,'' in \emph{European Conference on Computer Vision}, 2020.

\bibitem{de2019continual}
M.~De~Lange, R.~Aljundi, M.~Masana, S.~Parisot, X.~Jia, A.~Leonardis, G.~Slabaugh, and T.~Tuytelaars, ``{A continual learning survey: Defying forgetting in classification tasks},'' \emph{IEEE Transactions on Pattern Analysis and Machine Intelligence}, 2021.

\bibitem{parisi2019continual}
G.~I. Parisi, R.~Kemker, J.~L. Part, C.~Kanan, and S.~Wermter, ``{Continual lifelong learning with neural networks: A review},'' \emph{Neural Networks}, 2019.

\bibitem{mccloskey1989catastrophic}
M.~McCloskey and N.~J. Cohen, ``{Catastrophic interference in connectionist networks: The sequential learning problem},'' \emph{Psychology of learning and motivation}, 1989.

\bibitem{kirkpatrick2017overcoming}
J.~Kirkpatrick, R.~Pascanu, N.~Rabinowitz, J.~Veness, G.~Desjardins, A.~A. Rusu, K.~Milan, J.~Quan, T.~Ramalho, A.~Grabska-Barwinska \emph{et~al.}, ``{Overcoming catastrophic forgetting in neural networks},'' \emph{Proceedings of the National Academy of Sciences}, 2017.

\bibitem{zenke2017continual}
F.~Zenke, B.~Poole, and S.~Ganguli, ``{Continual learning through synaptic intelligence},'' in \emph{International Conference on Machine Learning}, 2017.

\bibitem{schwarz2018progress}
J.~Schwarz, W.~Czarnecki, J.~Luketina, A.~Grabska-Barwinska, Y.~W. Teh, R.~Pascanu, and R.~Hadsell, ``Progress \& compress: A scalable framework for continual learning,'' in \emph{International Conference on Machine Learning}, 2018.

\bibitem{mallya2018packnet}
A.~Mallya and S.~Lazebnik, ``{Packnet: Adding multiple tasks to a single network by iterative pruning},'' in \emph{Proceedings of the IEEE conference on Computer Vision and Pattern Recognition}, 2018.

\bibitem{robins1995catastrophic}
A.~Robins, ``{Catastrophic forgetting, rehearsal and pseudorehearsal},'' \emph{Connection Science}, 1995.

\bibitem{rebuffi2017icarl}
S.-A. Rebuffi, A.~Kolesnikov, G.~Sperl, and C.~H. Lampert, ``{iCaRL: Incremental classifier and representation learning},'' in \emph{Proceedings of the IEEE conference on Computer Vision and Pattern Recognition}, 2017.

\bibitem{buzzega2020dark}
P.~Buzzega, M.~Boschini, A.~Porrello, D.~Abati, and S.~Calderara, ``{Dark Experience for General Continual Learning: a Strong, Simple Baseline},'' in \emph{Advances in Neural Information Processing Systems}, 2020.

\bibitem{aljundi2019task}
R.~Aljundi, K.~Kelchtermans, and T.~Tuytelaars, ``Task-free continual learning,'' in \emph{Proceedings of the IEEE/CVF Conference on Computer Vision and Pattern Recognition}, 2019, pp. 11\,254--11\,263.

\bibitem{van2022three}
G.~M. van~de Ven, T.~Tuytelaars, and A.~S. Tolias, ``{Three types of incremental learning},'' \emph{Nature Machine Intelligence}, 2022.

\bibitem{mai2022online}
Z.~Mai, R.~Li, J.~Jeong, D.~Quispe, H.~Kim, and S.~Sanner, ``Online continual learning in image classification: An empirical survey,'' 2022.

\bibitem{ratcliff1990connectionist}
R.~Ratcliff, ``{Connectionist models of recognition memory: constraints imposed by learning and forgetting functions.}'' \emph{Psychological Review}, 1990.

\bibitem{aljundi2019gradient}
R.~Aljundi, M.~Lin, B.~Goujaud, and Y.~Bengio, ``{Gradient Based Sample Selection for Online Continual Learning},'' in \emph{Advances in Neural Information Processing Systems}, 2019.

\bibitem{chaudhry2021using}
A.~Chaudhry, A.~Gordo, P.~Dokania, P.~Torr, and D.~Lopez-Paz, ``Using hindsight to anchor past knowledge in continual learning,'' in \emph{Proceedings of the AAAI Conference on Artificial Intelligence}, 2021.

\bibitem{de2021continual}
M.~De~Lange and T.~Tuytelaars, ``Continual prototype evolution: Learning online from non-stationary data streams,'' in \emph{IEEE International Conference on Computer Vision}, 2021.

\bibitem{caccia2022new}
L.~Caccia, R.~Aljundi, N.~Asadi, T.~Tuytelaars, J.~Pineau, and E.~Belilovsky, ``{New Insights on Reducing Abrupt Representation Change in Online Continual Learning},'' in \emph{International Conference on Learning Representations Workshop}, 2022.

\bibitem{mai2021supervised}
Z.~Mai, R.~Li, H.~Kim, and S.~Sanner, ``Supervised contrastive replay: Revisiting the nearest class mean classifier in online class-incremental continual learning,'' in \emph{IEEE International Conference on Computer Vision and Pattern Recognition Workshops}, 2021.

\bibitem{guo2022online}
Y.~Guo, B.~Liu, and D.~Zhao, ``Online continual learning through mutual information maximization,'' in \emph{International Conference on Machine Learning}, 2022.

\bibitem{boschini2022transfer}
M.~Boschini, L.~Bonicelli, A.~Porrello, G.~Bellitto, M.~Pennisi, S.~Palazzo, C.~Spampinato, and S.~Calderara, ``Transfer without forgetting,'' in \emph{European Conference on Computer Vision}, 2022.

\bibitem{neuroai}
\BIBentryALTinterwordspacing
A.~Zador, S.~Escola, B.~Richards, B.~Ölveczky, Y.~Bengio, K.~Boahen, M.~Botvinick, D.~Chklovskii, A.~Churchland, C.~Clopath, J.~DiCarlo, S.~Ganguli, J.~Hawkins, K.~Koerding, A.~Koulakov, Y.~LeCun, T.~Lillicrap, A.~Marblestone, B.~Olshausen, A.~Pouget, C.~Savin, T.~Sejnowski, E.~Simoncelli, S.~Solla, D.~Sussillo, A.~S. Tolias, and D.~Tsao, ``Toward next-generation artificial intelligence: Catalyzing the neuroai revolution,'' \emph{arXiv preprint}, 2022. [Online]. Available: \url{https://arxiv.org/abs/2210.08340}
\BIBentrySTDinterwordspacing

\bibitem{ebrahimi2021remembering}
S.~Ebrahimi, S.~Petryk, A.~Gokul, W.~Gan, J.~E. Gonzalez, M.~Rohrbach, and T.~Darrell, ``Remembering for the right reasons: Explanations reduce catastrophic forgetting,'' \emph{Applied AI Letters}, 2021.

\bibitem{selvaraju2019grad}
R.~R. Selvaraju, M.~Cogswell, A.~Das, R.~Vedantam, D.~Parikh, and D.~Batra, ``{Grad-cam: Visual explanations from deep networks via gradient-based localization},'' \emph{International Journal of Computer Vision}, 2019.

\bibitem{saha2023saliency}
G.~Saha and K.~Roy, ``Saliency guided experience packing for replay in continual learning,'' in \emph{Proceedings of the IEEE/CVF Winter Conference on Applications of Computer Vision}, 2023, pp. 5273--5283.

\bibitem{bellitto2021hierarchical}
G.~Bellitto, F.~Proietto~Salanitri, S.~Palazzo, F.~Rundo, D.~Giordano, and C.~Spampinato, ``Hierarchical domain-adapted feature learning for video saliency prediction,'' \emph{International Journal of Computer Vision}, vol. 129, pp. 3216--3232, 2021.

\bibitem{wang2021spatio}
Z.~Wang, Z.~Liu, G.~Li, Y.~Wang, T.~Zhang, L.~Xu, and J.~Wang, ``Spatio-temporal self-attention network for video saliency prediction,'' \emph{IEEE Transactions on Multimedia}, 2021.

\bibitem{Hu_2023_WACV}
F.~Hu, S.~Palazzo, F.~P. Salanitri, G.~Bellitto, M.~Moradi, C.~Spampinato, and K.~McGuinness, ``Tinyhd: Efficient video saliency prediction with heterogeneous decoders using hierarchical maps distillation,'' in \emph{Proceedings of the IEEE/CVF Winter Conference on Applications of Computer Vision (WACV)}, 2023.

\bibitem{chaudhry2019tiny}
A.~Chaudhry, M.~Rohrbach, M.~Elhoseiny, T.~Ajanthan, P.~K. Dokania, P.~H. Torr, and M.~Ranzato, ``{On tiny episodic memories in continual learning},'' in \emph{International Conference on Machine Learning Workshop}, 2019.

\bibitem{ebrahimi2020adversarial}
S.~Ebrahimi, F.~Meier, R.~Calandra, T.~Darrell, and M.~Rohrbach, ``{Adversarial continual learning},'' in \emph{Proceedings of the European Conference on Computer Vision}, 2020.

\bibitem{derakhshani2021kernel}
M.~M. Derakhshani, X.~Zhen, L.~Shao, and C.~Snoek, ``{Kernel continual learning},'' in \emph{International Conference on Machine Learning}, 2021.

\bibitem{boschini2022class}
M.~Boschini, L.~Bonicelli, P.~Buzzega, A.~Porrello, and S.~Calderara, ``Class-incremental continual learning into the extended der-verse,'' \emph{IEEE Transactions on Pattern Analysis and Machine Intelligence}, 2022.

\bibitem{li2017learning}
Z.~Li and D.~Hoiem, ``{Learning without forgetting},'' \emph{IEEE Transactions on Pattern Analysis and Machine Intelligence}, 2017.

\bibitem{ewc}
J.~Kirkpatrick, R.~Pascanu, N.~Rabinowitz, J.~Veness, G.~Desjardins, A.~A. Rusu, K.~Milan, J.~Quan, T.~Ramalho, A.~Grabska-Barwinska, D.~Hassabis, C.~Clopath, D.~Kumaran, and R.~Hadsell, ``Overcoming catastrophic forgetting in neural networks,'' \emph{Proc. of the National Academy of Sciences}, 2017.

\bibitem{he2016deep}
K.~He, X.~Zhang, S.~Ren, and J.~Sun, ``{Deep residual learning for image recognition},'' in \emph{Proceedings of the IEEE conference on Computer Vision and Pattern Recognition}, 2016.

\bibitem{farquhar2018towards}
S.~Farquhar and Y.~Gal, ``{Towards Robust Evaluations of Continual Learning},'' in \emph{International Conference on Machine Learning Workshop}, 2018.

\bibitem{van2019three}
G.~M. van~de Ven and A.~S. Tolias, ``{Three continual learning scenarios},'' in \emph{Neural Information Processing Systems Workshops}, 2018.

\bibitem{wu2019large}
Y.~Wu, Y.~Chen, L.~Wang, Y.~Ye, Z.~Liu, Y.~Guo, and Y.~Fu, ``{Large scale incremental learning},'' in \emph{Proceedings of the IEEE conference on Computer Vision and Pattern Recognition}, 2019.

\bibitem{hou2019learning}
S.~Hou, X.~Pan, C.~C. Loy, Z.~Wang, and D.~Lin, ``{Learning a unified classifier incrementally via rebalancing},'' in \emph{Proceedings of the IEEE conference on Computer Vision and Pattern Recognition}, 2019.

\bibitem{zbontar2021barlow}
J.~Zbontar, L.~Jing, I.~Misra, Y.~LeCun, and S.~Deny, ``{Barlow twins: Self-supervised learning via redundancy reduction},'' in \emph{International Conference on Machine Learning}, 2021.

\bibitem{pmid19439676}
N.~Li, D.~D. Cox, D.~Zoccolan, and J.~J. DiCarlo, ``{{W}hat response properties do individual neurons need to underlie position and clutter "invariant" object recognition?}'' \emph{J Neurophysiol}, vol. 102, no.~1, pp. 360--376, Jul 2009.

\bibitem{spurious}
T.~Lesort, ``Continual feature selection: Spurious features in continual learning,'' 2022.

\bibitem{madry2018towards}
A.~Madry, A.~Makelov, L.~Schmidt, D.~Tsipras, and A.~Vladu, ``Towards deep learning models resistant to adversarial attacks,'' in \emph{International Conference on Learning Representations}, 2018.

\end{thebibliography}
\end{document}